\documentclass[10pt,twocolumn,letterpaper]{article}
\usepackage[pagenumbers]{cvpr}

\usepackage[skins]{tcolorbox}
\usepackage{fvextra}
\usepackage{stfloats}
\usepackage{balance}
\usepackage{threeparttable}
\makeatletter
\AddToHook{cmd/@BAlancecol/after}{%
  \setbox\@outputbox\vbox to \ht\@outputbox{%
    \protected@write\@auxout{}{\string\def\string\@LN@column{2}}%
    \unvbox\@outputbox
  }%
  \global\@colht=\oldvsize
  \@tempdima=\dimexpr\oldvsize-\ht\@leftcolumn-\dbltextfloatsep\relax
  \ifdim\@tempdima<\@dbltoproom
    \global\@dbltoproom=\@tempdima
  \fi
  \begingroup
    \def\f@depth{1sp}
    \@dbltopnum=0 
    \let\reserved@b\@deferlist
    \global\let\@deferlist\@empty
    \let\@elt\@sdblcolelt
    \reserved@b
  \endgroup
  \global\oldvsize=\@colht
}
\patchcmd{\@BAdblcol}{\@combinedblfloats}{%
  \@combinedblfloats
  \setbox\@outputbox\vbox to\textheight{\unvbox\@outputbox\vfil}%
}{}{}
\newcommand{\kmrolloutfloatsafterheading}{%
  \global\@dbltopnum=0
  \AddToHook{cmd/@dblfloatplacement/after}[km/rollout-order]{%
    \ifnum\c@page=\getpagerefnumber{sec:task_rollouts}\relax
      \global\@dbltopnum=0
    \fi
  }%
}
\makeatother
\newtcolorbox[auto counter]{kmpromptbox}[2][]{
  title={Template \thetcbcounter\quad #2},
  colback=black!2,colframe=black!55,
  colbacktitle=black!8,coltitle=black,
  fonttitle=\small\bfseries,boxrule=0.45pt,sharp corners,
  left=2mm,right=2mm,top=1mm,bottom=1mm,
  before skip=3pt,after skip=4pt,#1
}

\newcommand{\kmExtraVLACitations}{,zhang2026restacvla}

\usepackage{colortbl}
\definecolor{kmBestShade}{RGB}{224,224,255}
\definecolor{kmGroupShade}{RGB}{237,237,237}
\newcommand{\kmBest}[1]{\cellcolor{kmBestShade}\textbf{#1}}
\newcommand{\kmSecond}[1]{\underline{#1}}

\definecolor{cvprblue}{rgb}{0.21,0.49,0.74}
\usepackage[pagebackref,breaklinks,colorlinks,allcolors=cvprblue]{hyperref}

\newlength{\kmResourceIconHeight}
\newcommand{\kmGlobeIcon}{%
    \tikz[x=\kmResourceIconHeight, y=\kmResourceIconHeight,
          baseline=0pt, line width=.45pt, line cap=round]{
        \draw (0,.5) circle[radius=.46];
        \draw (0,.5) ellipse[x radius=.2, y radius=.46];
        \draw (-.46,.5) -- (.46,.5);
    }%
}
\newcommand{\kmGitHubIcon}{%
    \includegraphics[height=\kmResourceIconHeight]{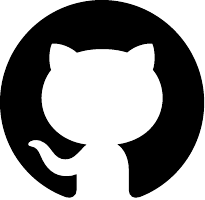}%
}
\newcommand{\kmResourceLink}[3]{%
    \begingroup
    \settoheight{\kmResourceIconHeight}{\textbf{H}}%
    #1\hspace{2pt}\href{#3}{\textbf{#2}}%
    \endgroup
}

\def\confName{CVPR}
\def\confYear{2027}

\title{KeyMemVLA: Subtask-Guided Key Memory for Long-Horizon Robotic Manipulation}
\author{%
    Yongwei Zhang\textsuperscript{1,2,*}\quad
    Tao Chen\textsuperscript{1,*}\quad
    Sicheng Ma\textsuperscript{2,3,*}\quad
    Yuanzhe Xing\textsuperscript{1}\quad
    Pengwei Zhang\textsuperscript{2,4,5}\\
    Yongen Zhao\textsuperscript{2,6}\quad
    Yaoqing Hu\textsuperscript{2}\quad
    Zhikun She\textsuperscript{1,7,\textdagger}\quad
    Ce Hao\textsuperscript{2,\textdagger}\\[3pt]
    {\small
        \textsuperscript{1}Beihang University\quad
        \textsuperscript{2}Zhongguancun Academy\quad
        \textsuperscript{3}Nanjing University
    }\\
    {\small
        \textsuperscript{4}Institute of Automation, Chinese Academy of Sciences\quad
        \textsuperscript{5}University of Chinese Academy of Sciences
    }\\
    {\small
        \textsuperscript{6}Tianjin University\quad
        \textsuperscript{7}Fujian Key Laboratory of Financial Information Processing, Putian University
    }\\[2pt]
    {\small
        \textsuperscript{*}Equal contribution\qquad
        \textsuperscript{\textdagger}Corresponding authors
    }\\[5pt]
    {\small
        \kmResourceLink{\kmGitHubIcon}{GitHub}{https://github.com/Yongwei-Zhang/KeyMemVLA/tree/main}\qquad
        \kmResourceLink{\kmGlobeIcon}{Project Page}{https://yongwei-zhang.github.io/keymemvla-project-page/}
    }%
}

\begin{document}
\maketitle
\begin{abstract}
Memory-dependent long-horizon manipulation requires vision-language-action (VLA) policies to infer task progress and recall evidence absent from current observations.
Under a limited memory budget, the challenge is to retain history that resolves stage ambiguity and supports subsequent decisions.
We present KeyMemVLA, a subtask-guided key memory framework based on progress indication and decision-evidence retention.
A two-level annotation process selects subtasks satisfying either criterion and labels their final frames as keyframes to supervise online memory writes.
A shared vision-language backbone jointly learns subtask, action, and keyframe prediction.
At inference, the policy first predicts the current subtask from the instruction, visual context, and memory, then uses this semantic condition to predict actions and future memory-write positions.
Actual observations are retained at the selected times during execution.
We introduce KMBench to evaluate active memory retention for decisions across subtasks.
Evaluations on RMBench-$M(n)$, RoboTwin-MeM, KMBench, and real-world manipulation tasks show higher average success rates than the compared baselines.
Ablation studies and representation analysis support the roles of key memory and explicit task-progress semantics.
These results highlight the value of coupling sparse key memory with online progress inference for history-dependent control.
\end{abstract}

\section{Introduction}
\label{sec:introduction}

\begin{figure*}[t]
    \centering
    \includegraphics[width=\textwidth]{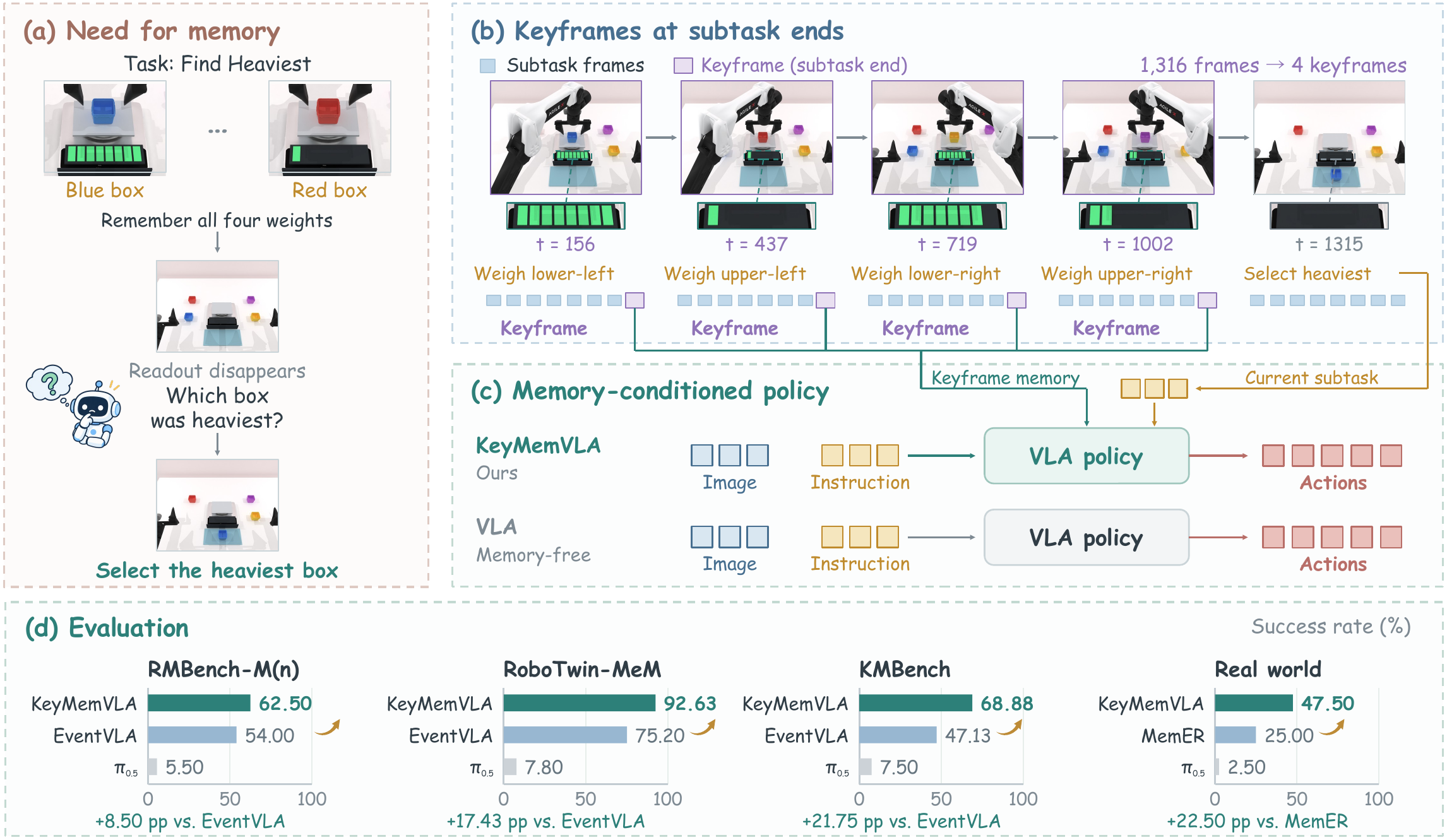}
    \caption{\textbf{Overview of KeyMemVLA.}
    \textbf{(a)} In \texttt{Find Heaviest}, selecting the heaviest box requires tracking weighing progress and recalling scale readouts after they disappear.
    \textbf{(b)} Four keyframes at the ends of key subtasks preserve progress cues and measurement evidence from a 1,316-frame trajectory.
    \textbf{(c)} Keyframe memory and online current-subtask prediction complement the instruction and visual context for action generation.
    \textbf{(d)} RMBench-$M(n)$ shows the w/o KPH variant jointly trained on all nine tasks; both KeyMemVLA and EventVLA are evaluated without keyframe memory.
    The remaining evaluations use the full KeyMemVLA policy.}
    \label{fig:keymemvla_overview}
\end{figure*}

\providecommand{\kmExtraVLACitations}{}
\edef\kmCiteVLAModels{\noexpand\cite{zitkovich2023rt2,kim2024openvla,black2025pi0,intelligence2025pi05\kmExtraVLACitations}}
Vision-language-action (VLA) models transfer pretrained visual and linguistic knowledge to robot control, enabling general instruction following and manipulation~\kmCiteVLAModels.
History-dependent long-horizon manipulation additionally requires information absent from the current observation.
In \texttt{Find Heaviest} (Fig.~\ref{fig:keymemvla_overview}(a)), a robot weighs four boxes, returns each to its original position, and selects the heaviest.
Similar workspace configurations recur after each measurement, while the scale readouts disappear.
The policy must distinguish which measurements have been completed and recall the evidence needed for comparison.
These are two distinct demands on memory: inferring task progress from similar observations and recovering historical evidence to choose an action even when the stage is known.

Existing approaches encode history through observation sequences, hierarchical memories, or compact learned states~\cite{guhur2022hiveformer,vanjani2026recat,tan2026memoact,zhou2026chronos,liu2026t2mem}.
Visual-memory methods preserve sampled video or selected perceptual evidence~\cite{shi2026memoryvla,shi2026memoryvlapp,yin2026simplememvla,zeng2026kemo,yang2026eventvla}, while memory-based planners and unified VLAs use history for subtask reasoning~\cite{sridhar2025memer,chen2026rmbench,osterberg2026unimem}.
Progress-aware retrieval, execution-state modeling, and completion checks further connect memory to task stages~\cite{chen2026keyframechaining,liu2026taskanchor,huang2026chainvla,liang2026tfp,zhao2026eventsync}.
These advances expose a consequential trade-off: extended histories increase processing costs, while compression and sparse selection risk losing brief cues needed later.
Under a finite budget, retained observations should supply progress cues or decision evidence beyond what the existing context provides.
The challenge we address is to make these task-dependent selection criteria explicit and coordinate memory retention with stage inference and action generation across subtasks.

We present \textbf{KeyMemVLA}, a subtask-guided key memory framework for history-dependent manipulation.
Our insight is that memory value depends on what a subtask establishes and what subsequent decisions require.
We therefore select history by two criteria: \emph{indicating task progress} and \emph{preserving evidence required for current or subsequent decisions}.
These criteria guide a two-level annotation process that selects key subtasks and informative end frames, forming sparse, temporally ordered key memory (Fig.~\ref{fig:keymemvla_overview}(b)).
Building on EventVLA's prediction of future memory-write positions~\cite{yang2026eventvla}, these annotations supervise a keyframe prediction head (KPH) that predicts when to retain actual observations during execution.
A shared vision-language backbone is jointly trained for subtask, action, and keyframe prediction.
At inference, it first predicts the current subtask from the instruction, observations, and memory, then conditions action generation and keyframe prediction on that subtask (Fig.~\ref{fig:keymemvla_overview}(c)).

We evaluate KeyMemVLA against reactive and memory-based policies on RMBench~\cite{chen2026rmbench}, RoboTwin-MeM~\cite{yang2026eventvla}, and \textbf{KMBench}, our eight-task suite testing active memory retention for subsequent decisions.
On RoboTwin-MeM and KMBench, KeyMemVLA achieves average task success rates of \textbf{92.63\% and 68.88\%}, improving over the compared EventVLA results by \textbf{17.43 and 21.75} percentage points, respectively.
On RMBench, the variant without KPH outperforms EventVLA on $M(n)$ tasks (Fig.~\ref{fig:keymemvla_overview}(d)).
Component ablations support the contributions of keyframe memory and online subtask prediction, while linear probes show improved decodability of temporal progress from the backbone representations used for action prediction.
Real-world comparisons with $\pi_{0.5}$ and MemER complement the simulation evaluation.

Our contributions are threefold:
\textbf{(i)} We develop subtask-guided key memory construction using progress indication and decision-evidence retention to select key subtasks and informative observations.
\textbf{(ii)} We build a shared-backbone VLA policy coupling sparse key memory with online subtask prediction, subtask-conditioned control, and learned memory updates.
\textbf{(iii)} We provide multi-benchmark evaluation, including our newly introduced KMBench comprising eight tasks across four memory categories, with component ablations and representation analysis supporting the framework's effectiveness and the roles of memory and explicit progress semantics.

\section{Related Work}
\label{sec:related_work}

\subsection{Memory-Augmented Robotic Manipulation}
\label{sec:related_memory}

Memory-augmented policies differ in how they represent and retain execution history.
History-aware transformers and recurrent policies encode observation--action sequences or latent states~\cite{guhur2022hiveformer,wu2024gr1,li2024roboflamingo,vanjani2026recat}.
Hierarchical memory, state-space models, and test-time fast-weight updates provide alternative mechanisms for maintaining long-term context~\cite{tan2026memoact,zhou2026chronos,liu2026t2mem}.
Visual memories retain feature banks, compress past views into tokens, or preserve sampled video histories~\cite{shi2026memoryvla,fang2025sam2act,wang2026nativemem,yin2026simplememvla}.

Sparse memory selects particular observations: KEMO uses kinematic events and visual filtering~\cite{zeng2026kemo}, EventVLA predicts future memory-write positions~\cite{yang2026eventvla}, and Keyframe-Chaining VLA retrieves selected frames according to execution progress~\cite{chen2026keyframechaining}.
Other representations combine short-term video with long-term text~\cite{mem2026multiscale}, integrate visual and textual memory with control~\cite{osterberg2026unimem}, or encode changes between visual states~\cite{na2026evomemvla}.
These approaches vary in temporal coverage, retained perceptual detail, and the cost of maintaining history.

\subsection{Task Progress Modeling for Robot Control}
\label{sec:related_reasoning}

Task-progress modeling distinguishes execution phases that may share similar observations.
Language-grounded planning and reasoning connect task semantics to control~\cite{ichter2023saycan,zawalski2025ecot}; memory-based planners additionally condition subtask selection on execution history~\cite{sridhar2025memer,chen2026rmbench}.
Explicit progress models use different state representations: TaskAnchor learns a stage coordinate, ChainVLA carries execution context across policy queries, and TFP maintains a dynamically updated progress belief~\cite{liu2026taskanchor,huang2026chainvla,liang2026tfp}.
Completion monitoring supports stage transitions: EventSync verifies subgoal completion before updating completed-subgoal history~\cite{zhao2026eventsync}.
Stage inference estimates the current execution phase, while completion verification establishes whether an operation has finished and its completion can be recorded.
Mem-0 selects the next subtask using a separate planner triggered by subtask completion~\cite{chen2026rmbench}, whereas our policy re-estimates the current subtask at each decision cycle.

\subsection{Benchmarks for Memory-Dependent Manipulation}
\label{sec:related_benchmarks}

Memory-focused benchmarks characterize different historical dependencies.
MemoryBench tests spatial and action-history recall~\cite{fang2025sam2act}, while MIKASA-Robo evaluates object, spatial, and sequential memory, as well as memory capacity~\cite{cherepanov2026mikasa}.
RMBench organizes tasks by memory complexity~\cite{chen2026rmbench}, and RoboMME distinguishes temporal, spatial, object, and procedural memory~\cite{dai2026robomme}.
Interaction-centric suites examine transient evidence and historical sequences: RoboTwin-MeM focuses on evidence acquired during interaction~\cite{yang2026eventvla}, while RoboMemArena covers composite manipulation and reports both task success and stage-level progress~\cite{lei2026robomemarena}.
Diagnostic extensions control information availability or interference: MIKASA-Robo-VLA combines hidden-cue tasks with reactive controls~\cite{cherepanov2026mikasavla}, and RoboMME-Interference inserts unrelated sessions into the supplied history~\cite{rathi2026robommeinterference}.

\section{Method}
\label{sec:method}

\subsection{Overview}
\label{sec:method_overview}

\begin{figure*}[t]
    \centering
    \includegraphics[width=\textwidth]{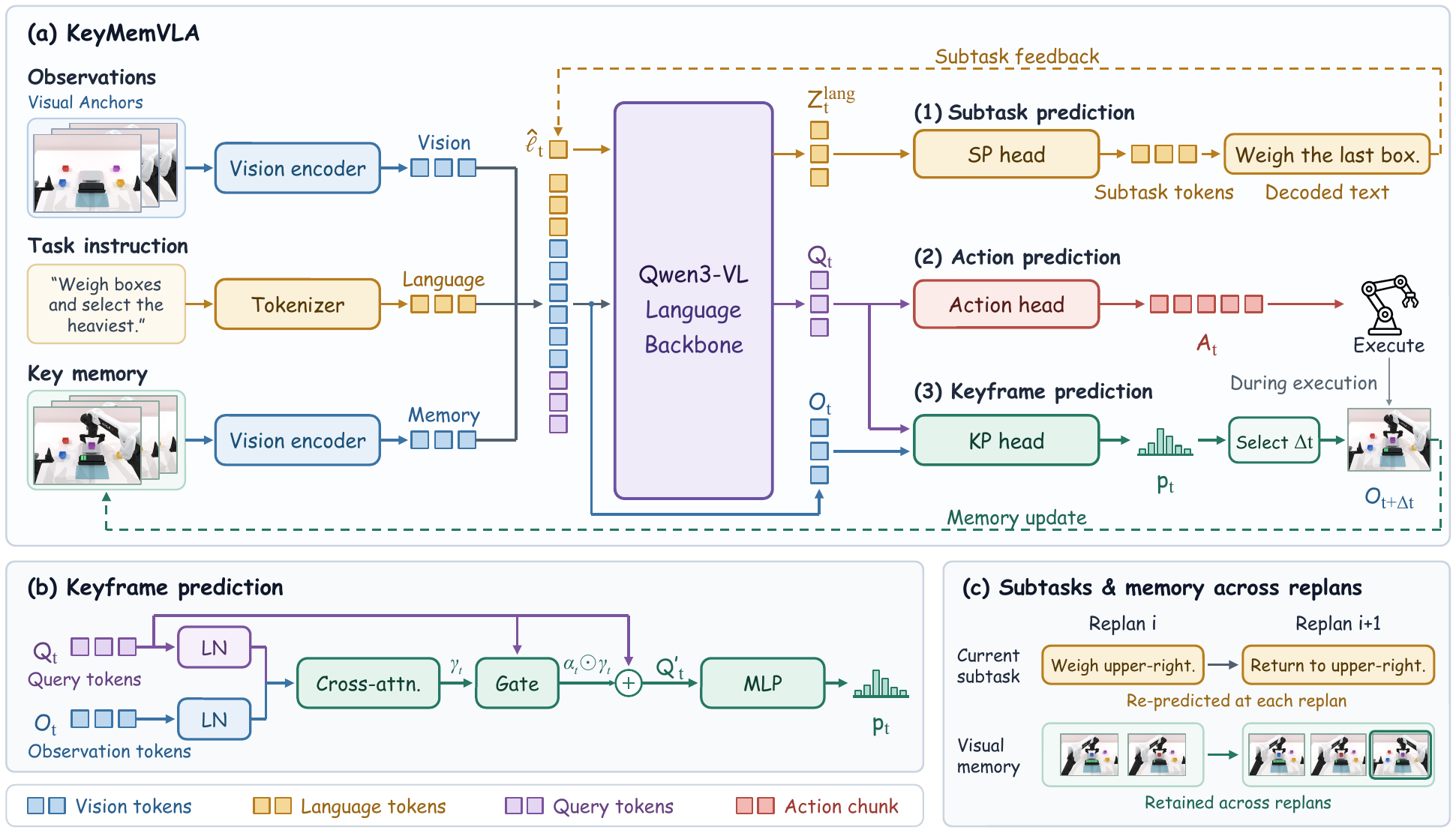}
    \caption{\textbf{KeyMemVLA framework.}
    \textbf{(a)} A shared vision-language backbone supports subtask, action, and keyframe prediction.
    Dashed paths show subtask feedback and memory updates during online inference.
    \textbf{(b)} Cross-attention uses $Q_t$ as queries and visual features $O_t$ from the current main-view image as keys and values.
    Gated residual fusion and an MLP head produce the keyframe scores $p_t$.
    \textbf{(c)} Current-subtask text is re-predicted at each replan, while visual key memory is retained across replans.}
    \label{fig:keymemvla_framework}
\end{figure*}

We define \emph{key memory} $M_t$ as task-relevant information selectively retained from execution history.
Selection follows two criteria.
\emph{Progress indication} retains cues that distinguish task stages with similar current observations.
\emph{Decision-evidence retention} preserves information needed for current or later decisions, even when the stage is known.

Let $H_t$ denote the complete observation--action history and $g$ the task instruction.
Following EventVLA's visual anchoring strategy~\cite{yang2026eventvla}, we group the initial observation, $K$ sampled recent observations, and the current observation into visual anchors $\mathrm{VA}_t=(o_0,o_{\tau_1},\ldots,o_{\tau_K},o_t)$, where $\tau_1<\cdots<\tau_K<t$.
The initial frame provides a stable reference and recent frames capture local temporal context, paralleling the functions of Mem-0's anchor and sliding memories~\cite{chen2026rmbench}.
In our implementation, current observations include all camera views, whereas historical anchors and keyframes use only the main view.
We aim to approximate the full-history optimal policy $\pi^\star$ with a policy $\bar\pi$ conditioned on these anchors and key memory:
\begin{equation}
    \begin{aligned}
        \pi^\star(\,\cdot\mid H_t,g)
        &\approx \bar{\pi}(\,\cdot\mid\mathrm{VA}_t,M_t,g),\\
        C(M_t)&\le B,
    \end{aligned}
    \label{eq:keymemvla_memory_objective}
\end{equation}
where $C(M_t)$ and $B$ denote the key-memory cost and budget.
Figure~\ref{fig:keymemvla_framework} summarizes how $g$, $\mathrm{VA}_t$, and $M_t$ condition a shared vision-language backbone.
The framework combines current-subtask prediction, subtask-conditioned action generation, and online memory updates to support decisions across task stages.

\subsection{Subtask-Guided Key Memory Construction}
\label{sec:key_memory_construction}

We construct key memory at two levels: key-subtask selection and keyframe extraction.
A subtask is selected if it satisfies at least one of the \emph{key memory} selection criteria in Sec.~\ref{sec:method_overview}: it provides an important cue to overall task progress, or it exposes evidence needed for current or later decisions.
In \texttt{Find Heaviest}, weighing a box marks measurement progress and yields evidence for the final comparison.
Returning it can be omitted from key-subtask selection because it adds neither necessary progress information nor comparison evidence.

For each key subtask, only its last frame is labeled as a keyframe.
Where needed, subtask boundaries are refined so that the final frame captures observable outcomes or transient evidence.
This frame marks a stage boundary and often preserves object attributes, spatial relations, or measurements needed later.
For example, the keyframe selected at the end of weighing shows the box together with its scale readout, recording progress and preserving evidence for later comparison (Fig.~\ref{fig:keymemvla_overview}(b)).
Let $\mathcal J$ index the selected key subtasks and $e_k$ denote the end time of subtask $k$.
The keyframe timestamps are
\begin{equation}
    \mathcal K=\{e_k\mid k\in\mathcal J\}.
    \label{eq:keyframe_annotation}
\end{equation}
These annotations supervise online memory-write prediction (Sec.~\ref{sec:training_pipeline}).
We use a semi-automatic annotation process.
An annotator first defines subtask segmentation and key subtask selection rules from representative demonstrations and encodes these rules in task-specific scripts.
These scripts are then applied to annotate the remaining demonstrations, with the last frame of each key subtask labeled as a keyframe.
Dataset-specific adaptations are detailed in Appendix~\ref{sec:annotation_pipeline}.

We consider three key memory modes.
\emph{ImageAndText} retains both keyframe images and their associated key-subtask descriptions.
\emph{Image} retains only the images, while \emph{Text} retains only the descriptions.
These modes specify only the content of $M_t$; all retain visual anchors and online current-subtask prediction.
The comparison of key memory modes favors \emph{Image} (Sec.~\ref{sec:memory_modality}), which we use for the main method and primary experiments.
Thus, $M_t=(o_{t'})_{t'\in\mathcal K_t}$, where $\mathcal K_t$ indexes past keyframes retained at time $t$, ordered chronologically.
Here, we use $C(M_t)=|M_t| \le B$ in Eq.~\eqref{eq:keymemvla_memory_objective}, where $|M_t|$ is the number of retained keyframes.

\subsection{Policy Architecture and Training}
\label{sec:training_pipeline}

Figure~\ref{fig:keymemvla_framework}(a) summarizes the policy architecture.
Our Qwen-OFT policy uses Qwen3-VL~\cite{bai2025qwen3vl} as its backbone and adopts action chunk prediction and continuous action regression from OFT~\cite{kim2025oft}.
Let $\mathcal C_t=(g,\mathrm{VA}_t,M_t)$ denote the input context and $Z_t$ the backbone hidden representations.
During training, only annotated keyframes strictly earlier than $t$ enter $M_t$.

The subtask prediction head (SPH) reuses the backbone's language generation head.
During training, the annotated current-subtask text $\ell_t$ supervises SPH through teacher forcing; at inference, $\ell_t$ is predicted by SPH.
The backbone then processes action-query tokens following $\ell_t$, yielding their representations in $Z_t$:
\begin{equation}
    Q_t=F_\theta^{\mathrm{qry}}(\mathcal C_t,\ell_t),
    \label{eq:action_queries}
\end{equation}
The VLM prompt templates are provided in Appendix~\ref{sec:vlm_prompt_design}.

A shared MLP action head maps $Q_t$ to a chunk $A_t$ of $L$ continuous actions.
For keyframe prediction, cross-attention uses $Q_t$ as queries and current main-view visual embeddings $O_t$ as keys and values.
A learned sigmoid gate controls the residual contribution of the attended visual features to $Q_t$.
The keyframe prediction head (KPH) reads the fused queries $Q'_t$ to predict per-position scores $p_t$, while the action head reads $Q_t$ (Fig.~\ref{fig:keymemvla_framework}(b)).
For $j=0,\ldots,L-1$, the score $p_{t,j}$ indicates whether to retain the observation at $t+j$.
We supervise these scores with soft targets based on temporal proximity to annotated keyframes:
\begin{equation}
    y_{t,j}=
    \begin{cases}
        \dfrac{1+\cos(\pi d_{t,j}/r)}{2}, & d_{t,j}\le r,\\
        0, & d_{t,j}>r,
    \end{cases}
    \label{eq:keyframe_soft_targets}
\end{equation}
where $d_{t,j}=\min_{t'\in\mathcal K}|t+j-t'|$ is the temporal distance to the nearest annotated keyframe and $r>0$ is the temporal radius.
We jointly optimize action regression, subtask prediction, and keyframe prediction using an L1 loss $\mathcal L_{\mathrm{act}}$, token cross-entropy $\mathcal L_{\mathrm{sph}}$, and binary cross-entropy $\mathcal L_{\mathrm{kph}}$:
\begin{equation}
    \mathcal{L}
    = \lambda_{\mathrm{act}}\mathcal{L}_{\mathrm{act}}
    + \lambda_{\mathrm{sph}}\mathcal{L}_{\mathrm{sph}}
    + \lambda_{\mathrm{kph}}\mathcal{L}_{\mathrm{kph}}.
    \label{eq:joint_training_loss}
\end{equation}
The weights $\lambda$ balance the objectives.
Module computations, additional soft-target details, and loss definitions are provided in Appendix~\ref{sec:architecture_training_details}.

\subsection{Online Inference}
\label{sec:online_inference}

Each episode starts with an empty key memory $M_0$.
At each decision cycle, inference uses the shared backbone in two stages (Fig.~\ref{fig:keymemvla_framework}(a)).
First, SPH generates the current subtask $\hat{\ell}_t$ from $\mathcal C_t$ autoregressively.
The predicted text is then fed back as the subtask response to obtain $Q_t$, from which the action head and KPH predict $A_t$ and $p_t$, respectively.

Based on the predicted keyframe scores, we select the highest-scoring future position in the action chunk if its score exceeds the write threshold.
When execution reaches the selected time, the captured image is added to key memory for subsequent decisions.
Temporal filtering suppresses repeated writes from nearby keyframe predictions, keeping the memory sparse.
Images are retained chronologically, with the oldest entries evicted as needed to maintain $C(M_t)\le B$.
Current-subtask text is re-predicted at each cycle, while visual key memory persists across cycles (Fig.~\ref{fig:keymemvla_framework}(c)).
Write filtering and execution details are provided in Appendices~\ref{sec:architecture_training_details} and~\ref{sec:real_world_deployment}.

\section{Experiments}
\label{sec:experiments}

\subsection{Simulation Experiments}
\label{sec:simulation_experiments}

\par
\phantomsection
\label{sec:rmbench}
\noindent\textbf{Evaluation on RMBench and the role of key memory.}
RMBench contains five $M(1)$ and four $M(n)$ tasks requiring one and multiple historical observations, respectively~\cite{chen2026rmbench}.
We compare two KeyMemVLA variants, each jointly trained on all nine tasks: one uses KPH and keyframe memory, while the other disables both. Both retain visual anchors and SPH.
Each policy is evaluated over 100 trials per task.
Keyframe annotations follow Appendix~\ref{sec:annotation_pipeline}.
Table~\ref{tab:rmbench_main} shows that enabling KPH and keyframe memory lowers overall success from 68.22\% to 64.56\%, with gains on some tasks and losses on others.
For some tasks, visual anchors may already provide sufficient cues, making additional keyframes redundant.

\begin{table}[t]
    \centering
    \begin{threeparttable}
    \caption{\textbf{RMBench success rates.}}
    \label{tab:rmbench_main}
    \small
    \setlength{\tabcolsep}{3pt}
    \renewcommand{\arraystretch}{0.96}
    \begin{tabular}{lccc}
        \toprule
        Method & $M(1)$ & $M(n)$ & Overall \\
        \midrule
        EventVLA (VA only)~\cite{yang2026eventvla} & \kmBest{79.00\%} & \kmSecond{54.00\%} & 67.80\% \\
        KeyMemVLA w/o KPH & 72.80\% & \kmBest{62.50\%} & 68.22\% \\
        KeyMemVLA & \kmSecond{76.60\%} & 49.50\% & 64.56\% \\
        \bottomrule
    \end{tabular}
    \begin{tablenotes}[flushleft]
        \footnotesize
        \item[] \textit{Note.} Both KeyMemVLA policies jointly train on all nine tasks and retain SPH.
        The w/o KPH variant disables keyframe prediction and memory; EventVLA uses VA only.
        Shaded bold and underlined values mark the best and second-best group averages.
    \end{tablenotes}
    \end{threeparttable}
\end{table}

Without KPH, overall success is comparable to EventVLA (VA only), with higher success on all four $M(n)$ tasks but a lower average success rate on $M(1)$ tasks.
Explicit current-subtask prediction may help distinguish repeated action stages, whereas EventVLA's richer initial and recent multi-view context may benefit $M(1)$ tasks.
These observations suggest that the value of additional memory depends on the visual context already available and the task's information requirements.
Full results, visual-context settings, and task-level analysis are in Appendix~\ref{sec:rmbench_additional_details}.

\par
\phantomsection
\label{sec:robotwin_mem}
\noindent\textbf{Evaluation on RoboTwin-MeM and intermediate evidence retention.}
RoboTwin-MeM, introduced by EventVLA~\cite{yang2026eventvla}, comprises eight bimanual tasks requiring retention of visual evidence that appears during interaction and later becomes unavailable.
We use the \emph{Image} mode defined in Sec.~\ref{sec:key_memory_construction}, retaining keyframe images without associated subtask descriptions.
We evaluate KeyMemVLA over 100 trials per task.
SPH still predicts the current subtask to condition action generation.
Key memory modes are compared in Sec.~\ref{sec:memory_modality}.

\begin{table*}[t]
    \centering
    \begin{threeparttable}
    \caption{\textbf{Task success rates on RoboTwin-MeM.}}
    \label{tab:robotwin_mem_main}
    \footnotesize
    \setlength{\tabcolsep}{1pt}
    \renewcommand{\arraystretch}{0.94}
    \begin{tabular*}{\textwidth}{@{\extracolsep{\fill}}lccccccccc@{}}
        \toprule
        & \multicolumn{1}{c}{$n=1$} & \multicolumn{1}{c}{$n=2$}
        & \multicolumn{2}{c}{$n=3$} & \multicolumn{3}{c}{$n=4$} & \multicolumn{1}{c}{$n=5$} & \\
        \cmidrule(lr){2-2}\cmidrule(lr){3-3}\cmidrule(lr){4-5}\cmidrule(lr){6-8}\cmidrule(lr){9-9}
        Method & \shortstack{Rearrange\\Blocks Hard} & \shortstack{Put Back\\Block Hard}
        & \shortstack{Pick Objects\\in Order} & \shortstack{Pick the\\Unhidden Block}
        & \shortstack{Cover\\Blocks Hard} & \shortstack{Find Seal\\and Seal Stamp}
        & \shortstack{Reproduce\\Route} & \shortstack{Press Button\\Keyframe} & Average \\
        \midrule
        \rowcolor{kmGroupShade}
        \multicolumn{10}{l}{$\blacktriangledown$ \textit{Non-Memory-Based Policies}} \\
        $\pi_{0.5}$~\cite{intelligence2025pi05} & 20\% & 19\% & 1\% & 14\% & 0\% & 8\% & 0\% & 0\% & 7.80\% \\
        Qwen-OFT~\cite{kim2025oft} & 3\% & 26\% & 0\% & 0\% & 0\% & 0\% & 0\% & 1\% & 3.80\% \\
        \rowcolor{kmGroupShade}
        \multicolumn{10}{l}{$\blacktriangledown$ \textit{Dual-System Memory-Based Vision-Language-Action Models}} \\
        MemER~\cite{sridhar2025memer} & 32\% & 4\% & 12\% & 2\% & 0\% & 26\% & 3\% & 5\% & 10.50\% \\
        Mem-0~\cite{chen2026rmbench} & 0\% & 0\% & 0\% & 0\% & 0\% & 0\% & 0\% & 0\% & 0.00\% \\
        \rowcolor{kmGroupShade}
        \multicolumn{10}{l}{$\blacktriangledown$ \textit{End-to-End Memory-Based Vision-Language-Action Models}} \\
        MemoryVLA~\cite{shi2026memoryvla} & 39\% & 0\% & 1\% & 9\% & 1\% & 11\% & 0\% & 25\% & 10.80\% \\
        EventVLA~\cite{yang2026eventvla} & \kmSecond{62\%} & \kmSecond{93\%} & \kmSecond{90\%} & \kmSecond{54\%} & \kmBest{94\%} & \kmSecond{63\%} & \kmSecond{98\%} & \kmSecond{48\%} & \kmSecond{75.20\%} \\
        \textbf{KeyMemVLA} & \kmBest{96\%} & \kmBest{100\%} & \kmBest{100\%} & \kmBest{100\%} & \kmSecond{91\%} & \kmBest{80\%} & \kmBest{100\%} & \kmBest{74\%} & \kmBest{92.63\%} \\
        \bottomrule
    \end{tabular*}
    \begin{tablenotes}[flushleft]
        \footnotesize
        \item[] \textit{Note.} Baselines follow EventVLA~\cite{yang2026eventvla}; $n$ denotes the maximum required keyframe count.
        Shaded bold and underlined entries denote the best and second-best distinct results, including ties.
    \end{tablenotes}
    \end{threeparttable}
\end{table*}

Table~\ref{tab:robotwin_mem_main} shows an average success rate of \textbf{92.63\%}, exceeding the full EventVLA model (VA+KEM; reported 75.20\%) by \textbf{17.43 pp}, with the best results on seven of eight tasks.
For \texttt{Pick the Unhidden Block}, which requires recalling briefly revealed colors, success increases from 54\% to 100\%.
These results support KeyMemVLA's effectiveness on manipulation tasks requiring intermediate visual evidence for later decisions; component contributions are examined in Sec.~\ref{sec:ablation_studies}.

\par
\phantomsection
\label{sec:kmbench}
\noindent\textbf{KMBench design and evaluation.}
We examine what information existing benchmarks require policies to remember.
On RMBench~\cite{chen2026rmbench}, our results suggest that initial observations and recent history already provide sufficient cues for many tasks (Appendix~\ref{sec:rmbench_additional_details}).
For example, the initial observation provides block colors and positions in \texttt{Cover Blocks}.
RoboTwin-MeM~\cite{yang2026eventvla} tests memory for transient cues such as the colors revealed in \texttt{Pick the Unhidden Block}.
However, four of its eight tasks acquire cues by lifting covers, with limited coverage of measurement comparison, calibration, and reasoning from pairwise test feedback.
We introduce KMBench to broaden memory evaluation through diverse task interactions.

\begin{figure*}[t]
    \centering
    \includegraphics[width=\textwidth]{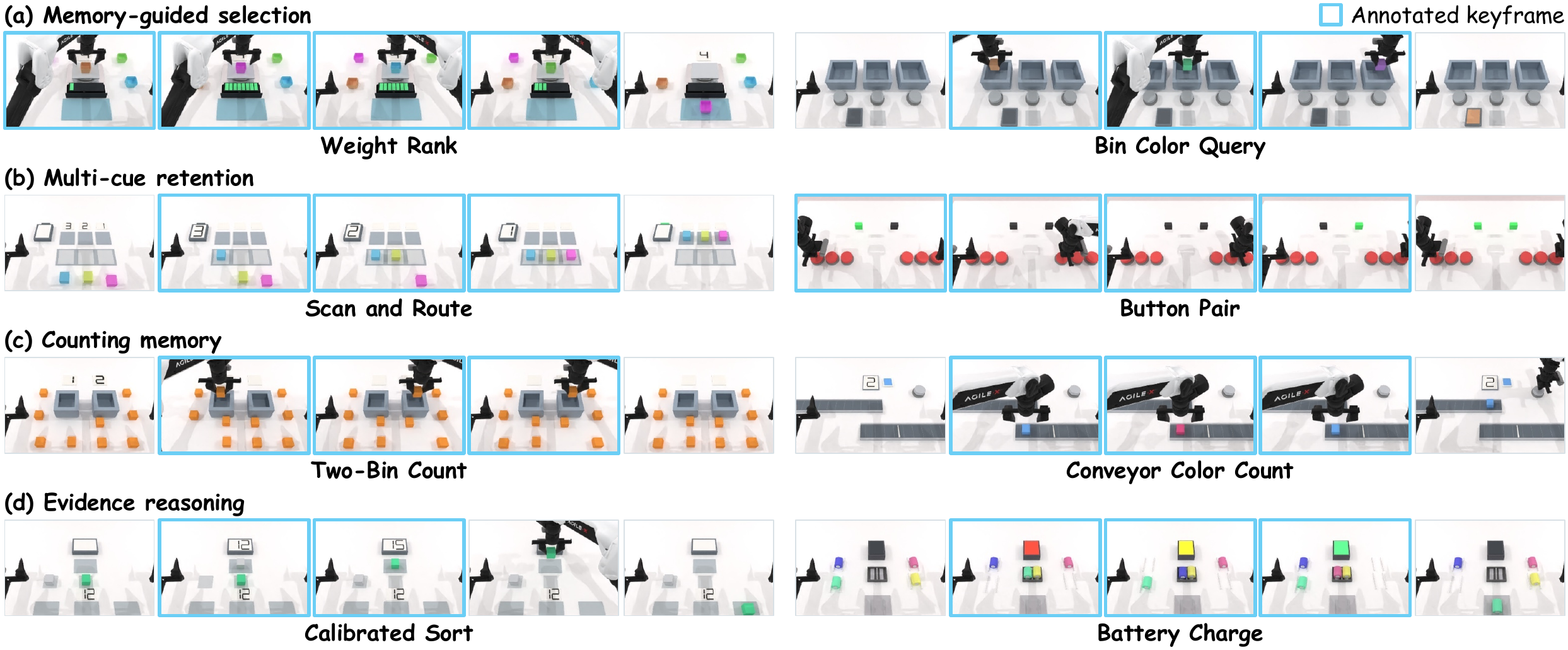}
    \caption{\textbf{KMBench task overview.}
    Eight tasks cover \textbf{(a)} memory-guided selection, \textbf{(b)} multi-cue retention, \textbf{(c)} counting memory, and \textbf{(d)} evidence reasoning.
    Each strip shows representative observations from one episode; cyan borders mark annotated keyframes.}
    \label{fig:kmbench_overview}
\end{figure*}

Figure~\ref{fig:kmbench_overview} illustrates eight bimanual tasks organized around four memory requirements.
\emph{Memory-guided selection} retains evidence associated with objects or locations for a later query.
\emph{Multi-cue retention} integrates associations acquired across exploration and execution stages.
\emph{Counting memory} tracks completed events whose totals are not visible in the current scene.
\emph{Evidence reasoning} combines earlier measurements or test feedback to support calibration and inference.
For example, the \emph{evidence reasoning} task \texttt{Battery Charge} evaluates whether feedback from different battery pairs can be combined to identify the lowest-charge battery.
Policies must actively retain task-relevant memory and its associations with objects, locations, or events to support subsequent decisions.
Task definitions and memory requirements are detailed in Appendix~\ref{sec:kmbench_additional_details}.

We collect 50 demonstrations per task (400 total) and evaluate each method over 100 trials per task, reporting the equal-weight average success rate.
Table~\ref{tab:kmbench_main} shows an average success rate of \textbf{68.88\%}, exceeding EventVLA's 47.13\% by \textbf{21.75 pp}, with the best results on six of eight tasks.
The two counting tasks show the largest gains over EventVLA: \texttt{Two-Bin Count} improves from 35\% to 97\%, and \texttt{Conveyor Color Count} from 10\% to 57\%.
\texttt{Battery Charge} also improves from 67\% to 91\%.
These results support the framework's effectiveness in tracking cumulative events and combining earlier test feedback for decisions across subtasks.
Component ablations are reported in Sec.~\ref{sec:ablation_studies}, with per-task results in Appendix~\ref{sec:ablation_probe_details}.

\begin{table*}[t]
    \centering
    \begin{threeparttable}
    \caption{\textbf{Task success rates on KMBench.}}
    \label{tab:kmbench_main}
    \footnotesize
    \setlength{\tabcolsep}{2pt}
    \renewcommand{\arraystretch}{0.94}
    \begin{tabular*}{\textwidth}{@{\extracolsep{\fill}}lccccccccc@{}}
        \toprule
        & \multicolumn{2}{c}{Memory-guided selection} & \multicolumn{2}{c}{Multi-cue retention}
        & \multicolumn{2}{c}{Counting memory} & \multicolumn{2}{c}{Evidence reasoning} & \\
        \cmidrule(lr){2-3}\cmidrule(lr){4-5}\cmidrule(lr){6-7}\cmidrule(lr){8-9}
        Method & \shortstack{Weight\\Rank} & \shortstack{Bin Color\\Query}
        & \shortstack{Scan and\\Route} & \shortstack{Button\\Pair}
        & \shortstack{Two-Bin\\Count} & \shortstack{Conveyor\\Color Count}
        & \shortstack{Calibrated\\Sort} & \shortstack{Battery\\Charge} & Average \\
        \midrule
        \rowcolor{kmGroupShade}
        \multicolumn{10}{l}{$\blacktriangledown$ \textit{Non-Memory-Based Policies}} \\
        ACT~\cite{zhao2023act} & 0\% & 0\% & 0\% & 12\% & 0\% & 3\% & 0\% & 0\% & 1.88\% \\
        DP~\cite{chi2023diffusion} & 0\% & 0\% & 0\% & 0\% & 8\% & 0\% & 0\% & 0\% & 1.00\% \\
        $\pi_{0.5}$~\cite{intelligence2025pi05} & 10\% & 3\% & 3\% & 4\% & 24\% & 1\% & 1\% & 14\% & 7.50\% \\
        \rowcolor{kmGroupShade}
        \multicolumn{10}{l}{$\blacktriangledown$ \textit{Memory-Based Vision-Language-Action Models}} \\
        MemER~\cite{sridhar2025memer} & \kmBest{29\%} & 22\% & 19\% & 4\% & 30\% & \kmSecond{13\%} & 39\% & 21\% & 22.13\% \\
        MemoryVLA~\cite{shi2026memoryvla} & 5\% & 0\% & 0\% & 12\% & 0\% & 0\% & 0\% & 9\% & 3.25\% \\
        EventVLA~\cite{yang2026eventvla} & 6\% & \kmBest{96\%} & \kmSecond{29\%} & \kmSecond{93\%} & \kmSecond{35\%} & 10\% & \kmSecond{41\%} & \kmSecond{67\%} & \kmSecond{47.13\%} \\
        \textbf{KeyMemVLA} & \kmSecond{26\%} & \kmSecond{90\%} & \kmBest{48\%} & \kmBest{97\%} & \kmBest{97\%} & \kmBest{57\%} & \kmBest{45\%} & \kmBest{91\%} & \kmBest{68.88\%} \\
        \bottomrule
    \end{tabular*}
    \begin{tablenotes}[flushleft]
        \footnotesize
        \item[] \textit{Note.} Task order and categories follow Fig.~\ref{fig:kmbench_overview}.
        Shaded bold and underlined values denote the best and second-best distinct results per column, including ties.
    \end{tablenotes}
    \end{threeparttable}
\end{table*}

\subsection{Real-World Experiments}
\label{sec:real_world}

\begin{figure*}[t]
    \centering
    \includegraphics[width=\textwidth]{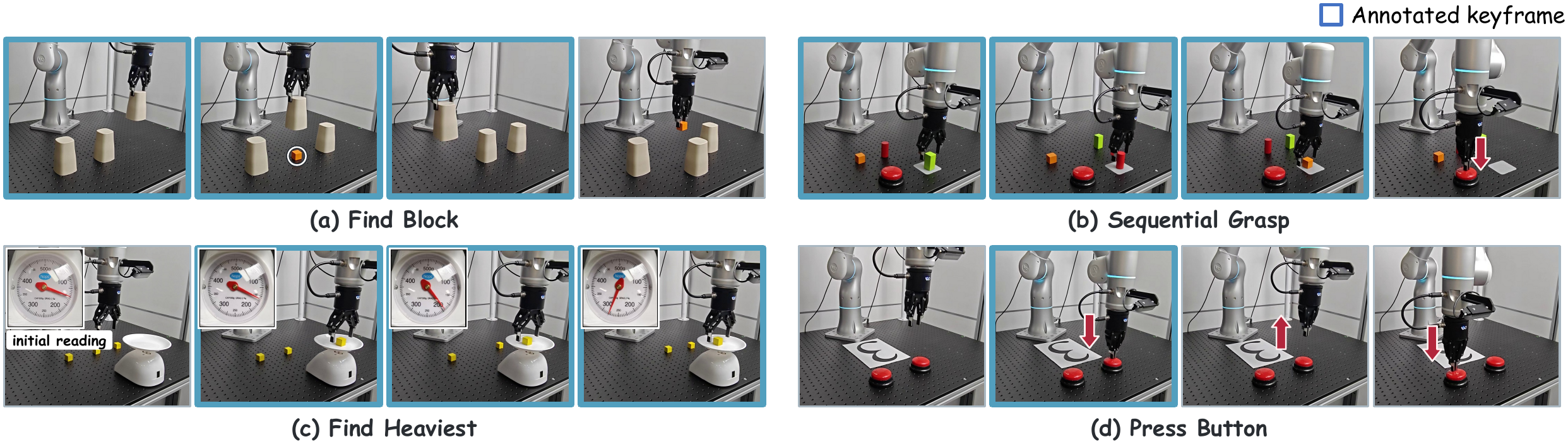}
    \caption{\textbf{Real-world task overview.}
    Representative observations from four tasks on the Flexiv Rizon4s: \textbf{(a)} \texttt{Find Block}, \textbf{(b)} \texttt{Sequential Grasp}, \textbf{(c)} \texttt{Find Heaviest}, and \textbf{(d)} \texttt{Press Button}.
    Blue borders mark annotated keyframes.}
    \label{fig:real_world_tasks}
\end{figure*}

We evaluate on a Flexiv Rizon4s robot with a gripper across four tasks (Fig.~\ref{fig:real_world_tasks}).
\texttt{Find Block} tests location recall after exploration; \texttt{Sequential Grasp} tests progress and order tracking when objects return to their initial positions; \texttt{Find Heaviest} tests comparison of earlier weight readouts; and \texttt{Press Button} tests cumulative counting.
Each task has 50 demonstrations (200 total).
We compare with the non-memory policy $\pi_{0.5}$~\cite{intelligence2025pi05} and memory-based MemER~\cite{sridhar2025memer}, using 10 trials per method and task.
Camera configurations and success criteria are detailed in Appendix~\ref{sec:real_world_evaluation_details}.

\begin{table}[t]
    \centering
    \begin{threeparttable}
    \caption{\textbf{Real-world task success rates.}}
    \label{tab:real_world_main}
    \small
    \setlength{\tabcolsep}{1.5pt}
    \renewcommand{\arraystretch}{0.94}
    \begin{tabular*}{\columnwidth}{@{\extracolsep{\fill}}lccccc@{}}
        \toprule
        Method & \shortstack{Find\\Block} & \shortstack{Sequential\\Grasp}
        & \shortstack{Find\\Heaviest} & \shortstack{Press\\Button} & Average \\
        \midrule
        $\pi_{0.5}$~\cite{intelligence2025pi05} & 0\% & 0\% & 0\% & 10\% & 2.5\% \\
        MemER~\cite{sridhar2025memer} & \kmSecond{20\%} & \kmSecond{40\%} & \kmSecond{20\%} & \kmSecond{20\%} & \kmSecond{25.0\%} \\
        \textbf{KeyMemVLA} & \kmBest{50\%} & \kmBest{50\%} & \kmBest{30\%} & \kmBest{60\%} & \kmBest{47.5\%} \\
        \bottomrule
    \end{tabular*}
    \begin{tablenotes}[flushleft]
        \footnotesize
        \item[] \textit{Note.} Each entry uses 10 trials; Average is computed over the four tasks.
        Shaded bold and underlined values denote the best and second-best results.
    \end{tablenotes}
    \end{threeparttable}
\end{table}

Table~\ref{tab:real_world_main} shows an average success rate of \textbf{47.5\%}, compared with 25.0\% for MemER and 2.5\% for $\pi_{0.5}$, with higher success on all four tasks.
\texttt{Find Block} improves from 20\% to 50\% and \texttt{Press Button} from 20\% to 60\% over MemER, supporting the framework's use of retained exploration evidence and cumulative count tracking.
We further compare inference efficiency on \texttt{Sequential Grasp} using a single NVIDIA A100 GPU (Appendix~\ref{sec:real_world_evaluation_details}).
Compared with MemER, KeyMemVLA more than halves mean inference latency (640\,ms vs.\ 1432\,ms) and reduces mean GPU memory usage (10.20\,GiB vs.\ 17.05\,GiB).

\subsection{Ablation and Representation Analysis}
\label{sec:ablation_studies}

\noindent\textbf{Component ablations.}
We compare the full \emph{Image} policy with two separately trained variants on RoboTwin-MeM and KMBench.
Removing KPH disables keyframe supervision, prediction, writes, and inputs; removing SPH disables subtask supervision, prediction, and text conditioning.
All variants retain visual anchors.
Across RoboTwin-MeM and KMBench, removing KPH reduces the average success rate by \textbf{61.00} and \textbf{47.13 pp}, respectively.
The corresponding reductions without SPH are \textbf{30.50} and \textbf{5.13 pp} (Table~\ref{tab:component_ablations}).

At the task level, removing KPH lowers success on \texttt{Put Back Block Hard} from 100\% to 38\% on RoboTwin-MeM and on \texttt{Bin Color Query} from 90\% to 0\% on KMBench.
These drops support the role of keyframe memory in preserving earlier visual evidence to complement visual anchors during subsequent decisions.
SPH also improves the average success rate on both benchmarks.
Removing SPH reduces success on \texttt{Find Seal and Seal Stamp} by 32 pp on RoboTwin-MeM.
The drops on KMBench's \texttt{Scan and Route} and \texttt{Button Pair} are 15 and 14 pp, respectively (Appendix~\ref{sec:ablation_probe_details}).
Explicit current-subtask prediction may help distinguish registration from delivery or exploration from final joint pressing, thereby guiding the policy to select actions appropriate to the current stage using retained evidence.

\begin{table}[t]
    \centering
    \begin{threeparttable}
    \caption{\textbf{Component ablations on RoboTwin-MeM and KMBench.}}
    \label{tab:component_ablations}
    \small
    \setlength{\tabcolsep}{4pt}
    \renewcommand{\arraystretch}{0.96}
    \begin{tabular*}{\columnwidth}{@{\extracolsep{\fill}}lcccc@{}}
        \toprule
        Model & KPH & SPH & Success & Drop (pp) \\
        \midrule
        \multicolumn{5}{@{}l}{\emph{RoboTwin-MeM}} \\
        KeyMemVLA & $\checkmark$ & $\checkmark$ & \kmBest{92.63\%} & --- \\
        \emph{w/o KPH} & $\times$ & $\checkmark$ & 31.63\% & 61.00 \\
        \emph{w/o SPH} & $\checkmark$ & $\times$ & 62.13\% & 30.50 \\
        \midrule
        \multicolumn{5}{@{}l}{\emph{KMBench}} \\
        KeyMemVLA & $\checkmark$ & $\checkmark$ & \kmBest{68.88\%} & --- \\
        \emph{w/o KPH} & $\times$ & $\checkmark$ & 21.75\% & 47.13 \\
        \emph{w/o SPH} & $\checkmark$ & $\times$ & 63.75\% & 5.13 \\
        \bottomrule
    \end{tabular*}
    \begin{tablenotes}[flushleft]
        \footnotesize
        \item[] \textit{Note.} Success is averaged over eight tasks per benchmark; Drop is relative to the corresponding full model, in percentage points.
    \end{tablenotes}
    \end{threeparttable}
\end{table}

\noindent\textbf{Progress representation.}
The success-rate drop without SPH highlights the importance of explicit progress prediction in long-horizon manipulation.
We therefore examine whether the full SPH pipeline makes temporal progress more linearly decodable from action representations.
On RoboTwin-MeM, we freeze the two VLA checkpoints and train a linear probe on each model's averaged action query representations to predict normalized frame progress, using paired visual contexts and fixed visible keyframes.
We measure progress prediction accuracy using mean absolute error (MAE) between predicted and ground-truth progress.
Across eight tasks, linear probes evaluated on 80 episodes held out from probe training yield average test MAEs of 0.115 without SPH and 0.088 with SPH, a \textbf{23.12\%} relative reduction.
The probe with SPH has lower error on seven of eight tasks, supporting improved linear decodability of temporal progress under the full SPH pipeline.
Figure~\ref{fig:sph_representation} shows two example tasks; the complete protocol and per-task results are in Appendix~\ref{sec:ablation_probe_details}.

\begin{figure}[t]
    \centering
    \includegraphics[width=0.9\columnwidth]{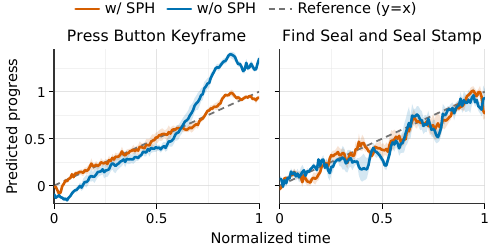}
    \caption{\textbf{Linear probing of temporal progress.}
    Average predictions for \texttt{Press Button Keyframe} and \texttt{Find Seal and Seal Stamp}, using 10 episodes per task held out from probe training.
    Shading denotes pointwise 95\% confidence intervals; the dashed line $y=x$ marks perfect agreement between predicted and ground-truth temporal progress.}
    \label{fig:sph_representation}
\end{figure}

\subsection{Key Memory Mode Analysis}
\label{sec:memory_modality}

We retain online SPH and compare the three historical key memory modes in Sec.~\ref{sec:key_memory_construction} on RoboTwin-MeM.
All modes use visual anchors and KPH.
Table~\ref{tab:memory_modality} reports \textbf{92.63\%} success for \emph{Image}, exceeding \emph{ImageAndText} (72.25\%) and \emph{Text} (53.88\%) by \textbf{20.38} and \textbf{38.75 pp}, respectively.

\begin{table}[t]
    \centering
    \begin{threeparttable}
    \caption{\textbf{Key memory modes on RoboTwin-MeM.}}
    \label{tab:memory_modality}
    \small
    \setlength{\tabcolsep}{4pt}
    \renewcommand{\arraystretch}{0.96}
    \begin{tabular}{lccc}
        \toprule
        Mode & Keyframes & Text history & Success \\
        \midrule
        \emph{Image} & $\checkmark$ & $\times$ & \kmBest{92.63\%} \\
        \emph{ImageAndText} & $\checkmark$ & $\checkmark$ & \kmSecond{72.25\%} \\
        \emph{Text} & $\times$ & $\checkmark$ & 53.88\% \\
        \bottomrule
    \end{tabular}
    \begin{tablenotes}[flushleft]
        \footnotesize
        \item[] \textit{Note.} Keyframes and Text history describe $M_t$; visual anchors, KPH, and online SPH are common to all modes.
        Shaded bold and underlined values denote the best and second-best results.
    \end{tablenotes}
    \end{threeparttable}
\end{table}

These differences may reflect both the information retained and how textual memory is updated.
Subtask descriptions may omit object attributes and spatial relations that keyframes preserve for later decisions, limiting \emph{Text} memory.
In modes that store text, the description is predicted when a write is planned and retained without a separate completion check.
Unverified subtask descriptions may bias progress inference and subsequent action prediction or conflict with actual keyframes in \emph{ImageAndText}.
Completion verification, as used in EventSync~\cite{zhao2026eventsync}, could therefore improve the reliability of textual key memory.
Under the current update procedure, we use \emph{Image} key memory together with online current-subtask prediction.
Appendix~\ref{sec:memory_mode_details} details memory construction, visibility rules, and per-task results.

\section{Conclusion}
\label{sec:conclusion}

We introduced KeyMemVLA, a subtask-guided framework that couples sparse key memory with online subtask prediction, action generation, and learned memory updates for long-horizon manipulation.
We also introduced KMBench to evaluate active memory retention for subsequent decisions across diverse task interactions.
Evaluations on RMBench-$M(n)$, RoboTwin-MeM, KMBench, and four real-world tasks show higher average success than the compared baselines.
Ablations support keyframe memory and the subtask prediction pipeline, while linear probes show improved temporal-progress decodability from the backbone representations used for action prediction.

\section*{LLM Usage Statement}
We used OpenAI Codex to assist with code debugging and bug fixes, language editing, manuscript consistency checks, figure layout revisions, and LaTeX preparation.
The authors take full responsibility for the final manuscript, annotations, implementation, and experimental results.

{
    \small
    \bibliographystyle{ieeenat_fullname}
    \bibliography{keymemvla}
}

\clearpage
\backrefsetup{disable}
\appendix
\maketitlesupplementary
\renewcommand{\floatpagefraction}{0.7}
\renewcommand{\dblfloatpagefraction}{0.7}
\section{Additional Method Details}
\label{sec:appendix_method}

\subsection{Script-Based Key-Subtask and Keyframe Annotation}
\label{sec:annotation_pipeline}

Section~\ref{sec:key_memory_construction} outlines the semi-automatic annotation process.
Here we specify interval constraints and dataset-specific adaptations.
Rules use segment labels, frame boundaries, or interaction events; additional demonstrations are inspected when target counts, exploration attempts, or execution branches differ.

\paragraph{KMBench and real-world tasks.}
For KMBench, collection scripts record basic action segments and their frame boundaries; task-specific rules merge adjacent segments into coherent stages of an operation.
For real-world demonstrations, boundaries are determined from collection records and video playback.
Using zero-based frame indices, we denote an episode of length $T$ by $\tau=((o_t,a_t))_{t=0}^{T-1}$, with subtask annotations
\begin{equation}
    \mathcal A(\tau)=(s_i)_{i=1}^{N},
    \qquad s_i=(\ell_i,[b_i,e_i]),
    \label{eq:subtask_intervals}
\end{equation}
where $\ell_i$ is a language description and $[b_i,e_i]$ is an inclusive frame interval.
The intervals are contiguous, non-overlapping, and cover the episode:
\begin{equation}
    \begin{aligned}
        b_1&=0,\qquad e_N=T-1,\qquad b_i\le e_i,\\
        b_{i+1}&=e_i+1\quad(1\le i<N).
    \end{aligned}
    \label{eq:subtask_coverage}
\end{equation}
A subtask is selected when it provides an important progress cue or exposes evidence required for current or later decisions, following the two criteria in Sec.~\ref{sec:key_memory_construction}.
Only the final frame of each selected subtask is marked as a keyframe, yielding $\mathcal K(\tau)=\{e_i\mid i\in\mathcal J\}$ as in Eq.~\ref{eq:keyframe_annotation}.
For example, the final frame of \emph{place the box on the scale and read its weight} shows both the box and the scale readout.
Returning the box is a subsequent subtask; once the box leaves the scale, the readout disappears.
Retaining the measurement frame supports the later comparison and selection of boxes.
Other subtask endpoints and the separately retained initial frame are excluded from $\mathcal K(\tau)$.

\paragraph{RMBench.}
RMBench already provides subtask annotations~\cite{chen2026rmbench}.
We inspect these annotations to determine which subtasks satisfy the key-memory criteria, then encode the selection rules in a task-specific script.
The script applies the rules to other episodes of the same task and labels the last frame of each selected subtask as a keyframe.

\paragraph{RoboTwin-MeM.}
RoboTwin-MeM provides both subtask annotations and keyframes capturing visual evidence that appears during interaction and later becomes unavailable~\cite{yang2026eventvla}.
Some keyframes precede their original subtask endpoints: there exist $t^*\in\mathcal K_{\mathrm{gt}}$ and an enclosing subtask $s_i$ such that $b_i\le t^*<e_i$.
We keep the provided keyframe set $\mathcal K_{\mathrm{gt}}$ unchanged and revise the relevant subtask boundaries and descriptions.
The revised annotations $\widetilde{\mathcal A}(\tau)=(\tilde s_i)_{i=1}^{N'}$, with $\tilde s_i=(\tilde\ell_i,[\tilde b_i,\tilde e_i])$, satisfy
\begin{equation}
    \mathcal K_{\mathrm{gt}}=\{\tilde e_i\mid i\in\widetilde{\mathcal J}\},
    \label{eq:aligned_keyframes}
\end{equation}
where $\widetilde{\mathcal J}$ indexes revised key subtasks.
Each original keyframe thus coincides with the endpoint of a revised key subtask, aligning the subtask decomposition with the terminal-frame rule while preserving the original keyframe timestamps.
Non-key subtasks remain part of the execution annotations and SPH supervision.

\subsection{Sequential Sampling for Multi-Task Training}
\label{sec:sequential_sampling}

The sampler balances tasks by repeating whole episodes and preserves temporal order within each batch slot.
Training uses annotated historical memory.

\paragraph{Valid starting points.}
For an episode $\tau$ of $T$ frames and action horizon $L$, a sample at $t$ supervises actions over $[t,t+L-1]$.
To ensure that the full action chunk lies within the episode, valid starting points satisfy $0\le t\le t_{\max}=T-L$.
For sampling interval $\Delta>1$, retain $t=0$, draw $u$ uniformly from $\{1,\ldots,\min(\Delta,t_{\max})\}$ when $t_{\max}\ge1$, and form
\begin{equation}
    \begin{aligned}
        \mathcal S_\tau^{(0)}=\{0\}\cup
        \{u+j\Delta\mid{}&j\in\mathbb Z_{\ge0},\\
                        &u+j\Delta\le t_{\max}\}.
    \end{aligned}
    \label{eq:sampler_anchors}
\end{equation}
We also consider $t_{\mathrm{tail}}=\min(T-1-\Delta,t_{\max})$ and add it if valid and absent, unless the preceding sample's action chunk already covers the episode end.
For $\Delta=1$, all valid starts are used.
The final set $\mathcal S_\tau$ is sorted chronologically.
We set the sampling interval $\Delta=L$, using 50 frames for simulation training and 30 frames for real-world training.

\paragraph{Task balancing.}
Let $\mathcal E_d$ be the episodes of task $d$ and $n_\tau=|\mathcal S_\tau|$.
Task totals and the common target are
\begin{equation}
    N_d=\sum_{\tau\in\mathcal E_d}n_\tau,\qquad
    N_{\mathrm{target}}=\max_d N_d.
    \label{eq:sampler_task_balance}
\end{equation}
We shuffle each task's episodes, repeat its full list $\lfloor N_{\mathrm{target}}/N_d\rfloor$ times, and append whole episodes until its total reaches the target.
The merged pool is shuffled before scheduling.
Repeated episodes share their starting points within an epoch.

\paragraph{Distributed batch slots.}
Episodes are assigned to the training process with the fewest accumulated samples, then sorted in descending order of $n_\tau$ and assigned to its least-loaded batch slot.
Each slot processes its episodes sequentially.
For stream length $n_{r,b}$ on process $r$ and slot $b$, all processes run
\begin{equation}
    N_{\mathrm{batch}}=\max_{r,b}n_{r,b}
    \label{eq:sampler_epoch_batches}
\end{equation}
batches per epoch.
Shorter streams cycle; empty slots reuse a nonempty stream.
Episode identifiers and boundary flags support memory updates and resets.

\subsection{VLM Prompt Design}
\label{sec:vlm_prompt_design}

All modes retain the task instruction, visual anchors, and current-subtask conditioning.
They differ in the representation of historical key-memory events (Table~\ref{tab:memory_prompt_inputs}).

\begin{table}[!htbp]
    \centering
    \begin{threeparttable}
    \caption{\textbf{Inputs for the three key memory modes.}}
    \label{tab:memory_prompt_inputs}
    \small
    \setlength{\tabcolsep}{3pt}
    \begin{tabular}{lccc}
        \toprule
        Mode & \shortstack{Visual\\anchors} & \shortstack{Historical\\images} & \shortstack{Historical\\text} \\
        \midrule
        \emph{Image} & $\checkmark$ & $\checkmark$ & $\times$ \\
        \emph{ImageAndText} & $\checkmark$ & $\checkmark$ & $\checkmark$ \\
        \emph{Text} & $\checkmark$ & $\times$ & $\checkmark$ \\
        \bottomrule
    \end{tabular}
    \begin{tablenotes}[flushleft]
        \footnotesize
        \item[] \textit{Note.} Historical images and text correspond to selected key-memory events; all modes retain visual anchors.
    \end{tablenotes}
    \end{threeparttable}
\end{table}

Templates~\ref{prompt:image}--\ref{prompt:text} show mode-specific inputs for $m$ historical key-memory events; Template~\ref{prompt:action} expands the shared \texttt{<action\_prompt>}.
All templates enable SPH, KPH, and image-group prefixes.
The backbone's chat template supplies role delimiters.
The \texttt{Current subtask:} field ends the user turn in \emph{Image} mode and begins the assistant turn in the other modes, matching their training formats.
For \emph{ImageAndText} and \emph{Text}, inference supplies \texttt{Current subtask:} as a fixed assistant prefix; the model generates only the subtask continuation.
The current subtask is annotated during training and generated at inference.
The \texttt{<q\_act>} token repeats $L$ times between the two \texttt{<action>} markers; the action head and KPH read these query positions.
Memory construction and visibility are detailed in Appendix~\ref{sec:memory_mode_details}.

\begin{kmpromptbox}[label=prompt:image]{\emph{Image}: keyframe images}
\begin{Verbatim}[fontsize=\small,breaklines=true,breakanywhere=true,breaksymbolleft={},breakindent=0pt,commandchars=\\\{\}]
\textbf{User:} Temporal observation images:
<temporal_observation_images>
Past keyframe images:
<keyframe_1> ... <keyframe_m>
Instruction: <task_instruction>
Current subtask:
\textbf{Assistant:} <current_subtask>
\textbf{User:} <action_prompt>
\end{Verbatim}
\end{kmpromptbox}
\begin{kmpromptbox}[label=prompt:image_text]{\emph{ImageAndText}: paired memory}
\begin{Verbatim}[fontsize=\small,breaklines=true,breakanywhere=true,breaksymbolleft={},breakindent=0pt,commandchars=\\\{\}]
\textbf{User:} Temporal observation images:
<temporal_observation_images>
Past keyframe images:
<keyframe_1> ... <keyframe_m>
Key subtasks corresponding to keyframes:
1. <key_subtask_1>
...
m. <key_subtask_m>
Instruction: <task_instruction>
Current subtask:
\textbf{Assistant:} <current_subtask>
\textbf{User:} <action_prompt>
\end{Verbatim}
\end{kmpromptbox}
\begin{kmpromptbox}[label=prompt:text]{\emph{Text}: subtask descriptions}
\begin{Verbatim}[fontsize=\small,breaklines=true,breakanywhere=true,breaksymbolleft={},breakindent=0pt,commandchars=\\\{\}]
\textbf{User:} Temporal observation images:
<temporal_observation_images>
Past key subtasks:
1. <key_subtask_1>
...
m. <key_subtask_m>
Instruction: <task_instruction>
Current subtask:
\textbf{Assistant:} <current_subtask>
\textbf{User:} <action_prompt>
\end{Verbatim}
\end{kmpromptbox}
\begin{kmpromptbox}[label=prompt:action]{Shared action prompt}
\begin{Verbatim}[fontsize=\small,breaklines=true,breakanywhere=true,breaksymbolleft={},breakindent=0pt,commandchars=\\\{\}]
Please predict the next <L> robot
actions and estimate which future
observation positions inside this
chunk are key events:
<action><q_act>...<q_act><action>.
\end{Verbatim}
\end{kmpromptbox}

\subsection{Model Architecture and Training Details}
\label{sec:architecture_training_details}

\paragraph{Architecture and initialization.}
The implementation builds on EventVLA~\cite{yang2026eventvla} and StarVLA's QwenOFT framework~\cite{community2026starvla}.
Simulation policies use Qwen3-VL-4B-Instruct; real-world policies use VLM weights from StarVLA/VLAct\_Qwen3\_Pretrain~\cite{yang2026vlact}.
SPH uses the backbone's language generation head.
The action head and KPH are initialized separately.

\paragraph{Hyperparameters.}
Tables~\ref{tab:robotwin_hyperparameters}--\ref{tab:real_world_deployment} summarize the dataset-specific model, training, and key-memory settings.
The RMBench configuration describes the w/o KPH policy.

\begin{table}[t]
        \caption{\textbf{RoboTwin-MeM hyperparameters.}}
        \label{tab:robotwin_hyperparameters}
        \centering
        \small
        \setlength{\tabcolsep}{2pt}
        \renewcommand{\arraystretch}{1.03}
        \begin{tabular}{@{}p{0.53\linewidth}>{\centering\arraybackslash}p{0.44\linewidth}@{}}
            \toprule
            \textbf{Configuration} & \textbf{Value} \\
            \midrule
            \multicolumn{2}{@{}l@{}}{\emph{Network architecture and basic training}} \\
            Base VLM & Qwen3-VL-4B-Instruct \\
            Image resolution & $256\times256$ \\
            Action head & Residual MLP (Qwen-OFT) \\
            Action horizon ($L$) & 50 \\
            Optimizer & AdamW \\
            Training steps & 100,000 \\
            Warm-up steps & 5,000 \\
            Base VLM / SPH learning rate & $1\times10^{-5}$ \\
            Action / KPH / connector LR & $1\times10^{-4}$ \\
            Per-device batch size & 4 \\
            Gradient accumulation steps & 1 \\
            \midrule
            \multicolumn{2}{@{}l@{}}{\emph{Key memory and supervision}} \\
            Key memory mode & Image \\
            Visual anchors & $o_0,\,o_{t-30},\,o_{t-15},\,o_t$ \\
            Training keyframe memory & Annotated (teacher) \\
            Memory capacity ($B$) & 5 \\
            Soft-target radius ($r$) & 8 steps \\
            Positive-class weight ($\beta$) & 7.0 \\
            Loss weights $(\lambda_{\mathrm{act}},\lambda_{\mathrm{sph}},\lambda_{\mathrm{kph}})$ & $(1.0,1.0,1.0)$ \\
            \bottomrule
        \end{tabular}
\end{table}

\begin{table}[t]
        \caption{\textbf{RMBench hyperparameters (w/o KPH).}}
        \label{tab:rmbench_hyperparameters}
        \centering
        \small
        \setlength{\tabcolsep}{2pt}
        \renewcommand{\arraystretch}{1.03}
        \begin{tabular}{@{}p{0.53\linewidth}>{\centering\arraybackslash}p{0.44\linewidth}@{}}
            \toprule
            \textbf{Configuration} & \textbf{Value} \\
            \midrule
            \multicolumn{2}{@{}l@{}}{\emph{Network architecture and basic training}} \\
            Base VLM & Qwen3-VL-4B-Instruct \\
            Image resolution & $256\times256$ \\
            Action head & Residual MLP (Qwen-OFT) \\
            Action horizon ($L$) & 50 \\
            Optimizer & AdamW \\
            Training steps & 100,000 \\
            Warm-up steps & 5,000 \\
            Base VLM / SPH learning rate & $1\times10^{-5}$ \\
            Action head learning rate & $1\times10^{-4}$ \\
            Per-device batch size & 4 \\
            Gradient accumulation steps & 1 \\
            \midrule
            \multicolumn{2}{@{}l@{}}{\emph{Subtask prediction and visual context}} \\
            Subtask prediction head (SPH) & Enabled \\
            KPH and keyframe memory & Disabled \\
            Visual anchors & $o_0,\,o_{t-30},\,o_{t-15},\,o_t$ \\
            Loss weights $(\lambda_{\mathrm{act}},\lambda_{\mathrm{sph}})$ & $(1.0,1.0)$ \\
            \bottomrule
        \end{tabular}
\end{table}

\begin{table}[t]
        \caption{\textbf{KMBench hyperparameters.}}
        \label{tab:kmbench_hyperparameters}
        \centering
        \small
        \setlength{\tabcolsep}{2pt}
        \renewcommand{\arraystretch}{1.03}
        \begin{tabular}{@{}p{0.53\linewidth}>{\centering\arraybackslash}p{0.44\linewidth}@{}}
            \toprule
            \textbf{Configuration} & \textbf{Value} \\
            \midrule
            \multicolumn{2}{@{}l@{}}{\emph{Network architecture and basic training}} \\
            Base VLM & Qwen3-VL-4B-Instruct \\
            Image resolution & $256\times256$ \\
            Action head & Residual MLP (Qwen-OFT) \\
            Action horizon ($L$) & 50 \\
            Optimizer & AdamW \\
            Training steps & 100,000 \\
            Warm-up steps & 5,000 \\
            Base VLM / SPH learning rate & $1\times10^{-5}$ \\
            Action / KPH / connector LR & $1\times10^{-4}$ \\
            Per-device batch size & 4 \\
            Gradient accumulation steps & 1 \\
            \midrule
            \multicolumn{2}{@{}l@{}}{\emph{Key memory and supervision}} \\
            Key memory mode & Image \\
            Visual anchors & $o_0,\,o_{t-30},\,o_{t-15},\,o_t$ \\
            Training keyframe memory & Annotated (teacher) \\
            Memory capacity ($B$) & 6 \\
            Soft-target radius ($r$) & 8 steps \\
            Positive-class weight ($\beta$) & 7.0 \\
            Loss weights $(\lambda_{\mathrm{act}},\lambda_{\mathrm{sph}},\lambda_{\mathrm{kph}})$ & $(1.0,1.0,1.0)$ \\
            \bottomrule
        \end{tabular}
\end{table}

\begin{table}[t]
        \caption{\textbf{Real-world hyperparameters.}}
        \label{tab:real_world_deployment}
        \centering
        \small
        \setlength{\tabcolsep}{2pt}
        \renewcommand{\arraystretch}{1.03}
        \begin{tabular}{@{}p{0.53\linewidth}>{\centering\arraybackslash}p{0.44\linewidth}@{}}
            \toprule
            \textbf{Configuration} & \textbf{Value} \\
            \midrule
            \multicolumn{2}{@{}l@{}}{\emph{Network architecture and basic training}} \\
            Base VLM & \shortstack{Qwen3-VL-4B\\(VLAct-pretrained)} \\
            Image resolution & $256\times256$ \\
            Action head & Residual MLP (Qwen-OFT) \\
            Action horizon ($L$) & 30 \\
            Optimizer & AdamW \\
            Training steps & 60,000 \\
            Warm-up steps & 2,000 \\
            Base VLM / SPH learning rate & $5\times10^{-5}$ \\
            Action / KPH / connector LR & $5\times10^{-5}$ \\
            Per-device batch size & 4 \\
            Gradient accumulation steps & 1 \\
            \midrule
            \multicolumn{2}{@{}l@{}}{\emph{Key memory and supervision}} \\
            Key memory mode & Image \\
            Visual anchors & $o_0,\,o_{t-30},\,o_{t-15},\,o_t$ \\
            Training keyframe memory & Annotated (teacher) \\
            Memory capacity ($B$) & 5 \\
            Soft-target radius ($r$) & 8 steps \\
            Positive-class weight ($\beta$) & 7.0 \\
            Loss weights $(\lambda_{\mathrm{act}},\lambda_{\mathrm{sph}},\lambda_{\mathrm{kph}})$ & $(1.0,1.0,1.0)$ \\
            \bottomrule
        \end{tabular}
\end{table}

\paragraph{Subtask prediction and action queries.}
The decision context is $\mathcal C_t=(g,\mathrm{VA}_t,M_t)$, and $Z_t$ denotes the shared backbone's final hidden representations.
SPH uses the backbone's existing language generation head to predict the current subtask, including subtasks not selected for key memory.
For a description $\ell_t$ containing $N_t$ tokens $w_{t,u}$, the prediction factorizes as
\begin{equation}
    P_\theta(\ell_t\mid\mathcal C_t)
    =\prod_{u=1}^{N_t}
    P_\theta(w_{t,u}\mid\mathcal C_t,w_{t,<u}).
    \label{eq:subtask_prediction}
\end{equation}
Training uses teacher forcing with annotated subtask tokens, as described in Sec.~\ref{sec:training_pipeline}.
The action-query representations $Q_t\in\mathbb R^{L\times d}$ follow Eq.~\eqref{eq:action_queries}, where $L$ is the action horizon and $d$ the backbone hidden dimension.
They are taken from the action-query token positions in $Z_t$.

\paragraph{Action generation.}
A shared residual MLP maps each query row $q_{t,j}$ to a continuous action:
\begin{equation}
    \begin{aligned}
        a_{t+j}&=h_{\mathrm{act}}(q_{t,j}),\quad j=0,\ldots,L-1,\\
        A_t&=(a_t,\ldots,a_{t+L-1})\in\mathbb R^{L\times d_a},
    \end{aligned}
    \label{eq:action_generation}
\end{equation}
where $d_a$ is the robot action dimension.
All $L$ actions are regressed together from $Q_t$, conditioned on the subtask response, task instruction, and visual context.

\paragraph{Keyframe prediction.}
KPH combines the action-query representations with visual-token embeddings $O_t\in\mathbb R^{n_v\times d}$ from the current main-view image.
As shown in Fig.~\ref{fig:keymemvla_framework}(b), its connector applies independently normalized cross-attention and gated residual fusion:
\begin{equation}
    \begin{aligned}
        \gamma_t&=\operatorname{CA}(\operatorname{LN}_q(Q_t),\operatorname{LN}_o(O_t)),\\
        \alpha_t&=\sigma([Q_t;\gamma_t]w_g+b_g),\\
        Q'_t&=Q_t+\alpha_t\odot\gamma_t,\\
        p_{t,j}&=\sigma\bigl(h_{\mathrm{kph}}(q'_{t,j})\bigr).
    \end{aligned}
    \label{eq:keyframe_prediction}
\end{equation}
Here, $\operatorname{CA}$ is multi-head cross-attention, $[\,;\,]$ concatenates features, and $\sigma$ is sigmoid.
The per-query scalar gate is broadcast over feature dimensions.
The MLP $h_{\mathrm{kph}}$ uses layer normalization, a GELU hidden layer, and a scalar output.
KPH reads $Q'_t$; the action head reads $Q_t$.

\paragraph{Temporal soft targets.}
The soft targets follow Eq.~\eqref{eq:keyframe_soft_targets}.
Distances use all trajectory annotations; an empty annotation set gives zero targets.
Thus, a keyframe just outside the current action chunk can still contribute to its soft targets.

\paragraph{Loss definitions.}
The three supervised losses are
\begin{equation}
    \begin{aligned}
        \mathcal L_{\mathrm{act}}
        &=\mathbb E\!\left[\frac{\|a_{t+j}-a^\star_{t+j}\|_1}{d_a}\right],\\
        \mathcal L_{\mathrm{sph}}
        &=-\mathbb E\!\left[\log P_\theta(w^\star_{t,u}\mid\mathcal C_t,w^\star_{t,<u})\right],\\
        \mathcal L_{\mathrm{kph}}
        &=-\mathbb E\!\left[\beta y_{t,j}\log p_{t,j}\right.\\
        &\hspace{3.6em}\left.{}+(1-y_{t,j})\log(1-p_{t,j})\right].
    \end{aligned}
    \label{eq:supervised_policy_losses}
\end{equation}
Superscript $\star$ denotes annotations, and $w_{t,u}$ is subtask token $u$; expectations average over examples and valid positions.
The positive-class weight is $\beta$; Eq.~\ref{eq:joint_training_loss} combines the objectives.

\paragraph{Online write scheduling.}
The procedure in Sec.~\ref{sec:online_inference} selects the future peak $j^\star=\operatorname*{arg\,max}_{1\le j<L}p_{t,j}$.
A write is armed only when $p_{t,j^\star}>\eta$ and the selected time lies within the executed action prefix.
Temporal filtering suppresses nearby repeated write proposals.
If replanning replaces the unexecuted portion of an action chunk, pending writes associated with that portion are canceled.
Write thresholds and memory capacities follow the dataset-specific configurations above.

\subsection{Real-World Inference and Execution}
\label{sec:real_world_deployment}

To implement online inference and memory updates on the real robot, we use the StarVLA WebSocket interface~\cite{community2026starvla} to connect a policy server running on one NVIDIA A100 GPU with a client responsible for observation acquisition, action execution, and memory management.
Visual anchors include the initial main-view image, historical main-view images at offsets $-30$ and $-15$, and current images from all three cameras; missing historical observations are replaced with the initial image.
At each policy query, the server first predicts the current subtask, followed by action and keyframe prediction.
As specified in Table~\ref{tab:real_world_deployment}, the policy predicts an action chunk of length 30, while the client executes at most 16 actions from each chunk.

\section{Benchmarks and Additional Experimental Details}
\label{sec:appendix_experiments}

\subsection{KMBench Task Definitions}
\label{sec:kmbench_additional_details}

Section~\ref{sec:kmbench} outlines the four memory requirements.
Here we specify task rules, evidence visibility, and conditions for the final decision.
Table~\ref{tab:kmbench_protocol} summarizes the KMBench setup and evaluation protocol.
Demonstration statistics are listed in Table~\ref{tab:kmbench_tasks}; all demonstrations include subtask and keyframe annotations.
Three $320\times240$ camera streams, proprioceptive states, and actions are recorded synchronously at 15\,Hz.

\begin{table}[!htbp]
    \centering
    \caption{\textbf{KMBench setup and evaluation protocol.}}
    \label{tab:kmbench_protocol}
    \small
    \setlength{\tabcolsep}{2pt}
    \renewcommand{\arraystretch}{1.03}
    \begin{tabular}{@{}>{\raggedright\arraybackslash}p{0.43\linewidth}>{\raggedright\arraybackslash}p{0.54\linewidth}@{}}
        \toprule
        Item & Setting \\
        \midrule
        Simulation environment & RoboTwin 2.0 \\
        Task suite & 8 bimanual tasks \\
        Memory categories & 4 \\
        Training demonstrations & 50 per task (400 total) \\
        Evaluation trials & 100 per method per task \\
        Per-task metric & Task success rate (\%) \\
        Overall metric & Equal-weight average across eight tasks \\
        \bottomrule
    \end{tabular}
\end{table}

\begin{table}[!htbp]
    \centering
    \begin{threeparttable}
    \caption{\textbf{KMBench training demonstrations.}}
    \label{tab:kmbench_tasks}
    \small
    \setlength{\tabcolsep}{0pt}
    \begin{tabular}{@{}>{\raggedright\arraybackslash}p{0.44\columnwidth}>{\centering\arraybackslash}p{0.28\columnwidth}>{\centering\arraybackslash}p{0.28\columnwidth}@{}}
        \toprule
        Task & Demonstrations & Frames \\
        \midrule
        Weight Rank & 50 & 70,596 \\
        Bin Color Query & 50 & 27,663 \\
        Scan and Route & 50 & 46,677 \\
        Button Pair & 50 & 23,693 \\
        Two-Bin Count & 50 & 39,137 \\
        Conveyor Color Count & 50 & 50,153 \\
        Calibrated Sort & 50 & 32,649 \\
        Battery Charge & 50 & 76,239 \\
        \bottomrule
    \end{tabular}
    \begin{tablenotes}[flushleft]
        \footnotesize
        \item[] \textit{Note.} All datasets are sampled at 15\,Hz.
    \end{tablenotes}
    \end{threeparttable}
\end{table}

\subsubsection{Memory-Guided Selection}

\paragraph{\texttt{Weight Rank}.}
Four boxes have randomized colors, weights, and positions.
During weighing, a display specifies a target rank $\rho\in\{1,2,3,4\}$.
The goal is to release the $\rho$-th lightest box on the central pad.
Demonstrations weigh all four boxes and return each to its original location; the readout disappears when the box leaves the scale.
This sequence requires tracking completed measurements and associating earlier readouts with the corresponding boxes for comparison.

\paragraph{\texttt{Bin Color Query}.}
The robot deposits three differently colored blocks into the left, center, and right bins in that order.
After each deposit, the block is removed before the next block is supplied.
Once all three blocks have been deposited and removed, a display presents a target color.
The robot then presses the button in front of the bin that received that color.
Because the deposited blocks are no longer visible at query time, the policy must retain the color-to-bin associations established during the transfers.

\subsubsection{Multi-Cue Retention}

\paragraph{\texttt{Scan and Route}.}
The robot registers three items and delivers them to designated stations.
The identifiers of the three stations are shown only in the initial frame.
Releasing an item on its registration pad reveals its destination identifier; registration is confirmed after the readout remains visible for one second, and the display then clears.
All three registrations must be complete and the display clear before delivery begins.
The robot then delivers and releases all three items at their assigned stations.
Routing therefore requires retaining the station-identifier-to-location and item-to-destination associations, together with delivery progress.

\paragraph{\texttt{Button Pair}.}
Each side of the workspace contains three candidate buttons, exactly one of which activates a light.
The light turns off when the button is released.
The goal is to activate both lights simultaneously by pressing the working button on each side.
Demonstrations explore the two sides and undergo resets before the final joint press, requiring the working-button locations to be retained across resets.

\subsubsection{Counting Memory}

\paragraph{\texttt{Two-Bin Count}.}
The robot deposits a specified number of orange blocks into each of two bins.
The left and right target counts, each ranging from 1 to 3, are shown only in the initial frame.
Deposited blocks are removed from the scene, while spare blocks are replenished on the table; the visible supply therefore does not reveal how many blocks each bin has received.
A deposit is counted when a released block enters a bin.
The policy must track completed deposits separately for the two bins and leave surplus blocks on the table.

\paragraph{\texttt{Conveyor Color Count}.}
Blocks of the displayed target color and distractor colors arrive in sequence.
The robot transfers them to an output conveyor in arrival order, returning the operating arm home after each transfer.
Only completed exits of target-color blocks count toward the displayed target, which ranges from 2 to 5.
Once this count is reached, the robot must finish the current transfer and arm return, then press the completion button without handling another block.
Because exited blocks disappear, the policy must track target-color exits separately from the total number of transfers.

\subsubsection{Evidence Reasoning}

\paragraph{\texttt{Calibrated Sort}.}
A measurement platform has an unknown additive offset that remains fixed within an episode.
The robot first measures a reference block with a known true value of 10, then a test block.
Each readout disappears when its object is removed.
Given readouts $r_{\mathrm{ref}}$ and $r_{\mathrm{test}}$, the calibrated test value is
\begin{equation}
    v=r_{\mathrm{test}}-r_{\mathrm{ref}}+10.
    \label{eq:kmbench_calibrated_value}
\end{equation}
After both valid readouts have been obtained and the display has cleared, the test block must be released on the left, center, or right output pad depending on whether $v$ is below, equal to, or above the displayed threshold.
The test block can move directly from the measurement platform to its output pad.
The decision combines the earlier reference readout with the test readout to remove their common offset.

\paragraph{\texttt{Battery Charge}.}
Four batteries have charges $\{1,2,2,3\}$, with the charge assignment randomized across episodes.
The robot must identify the lowest-charge battery and place it on the central pad.
A tester reports the sum of the charges of two selected batteries through a colored light: red for a sum of 3, yellow for 4, and green for at least 5.
The instruction specifies the charge multiset and light meanings; demonstrations use the lower-right battery as a fixed partner and test it with each of the other three.
Changing the pair removes or overwrites the previous feedback.
The policy associates these responses with the tested batteries and combines them with the known charge multiset to identify the lowest-charge battery.

\subsection{Real-World Task Definitions}
\label{sec:real_world_details}

Table~\ref{tab:real_world_protocol} summarizes the real-world setup and evaluation protocol.
Demonstration statistics are listed in Table~\ref{tab:real_world_demonstrations}; all datasets are sampled at 15\,Hz.
For \texttt{Press Button}, each target count has 10 demonstrations.

\begin{table}[!htbp]
    \centering
    \caption{\textbf{Real-world setup and evaluation protocol.}}
    \label{tab:real_world_protocol}
    \small
    \setlength{\tabcolsep}{2pt}
    \renewcommand{\arraystretch}{1.03}
    \begin{tabular}{@{}>{\raggedright\arraybackslash}p{0.43\linewidth}>{\raggedright\arraybackslash}p{0.54\linewidth}@{}}
        \toprule
        Item & Setting \\
        \midrule
        Robot platform & Flexiv Rizon4s with Flexiv-GN01 gripper \\
        Task suite & 4 real-world tasks \\
        Training demonstrations & 50 per task (200 total) \\
        Evaluation trials & 10 per method per task \\
        Per-task metrics & Task success rate (TSR) and subtask success rate (SSR), in \% \\
        Overall metric & Equal-weight average TSR across four tasks \\
        \bottomrule
    \end{tabular}
\end{table}

\begin{table}[!htbp]
    \centering
    \begin{threeparttable}
    \caption{\textbf{Real-world training demonstrations.}}
    \label{tab:real_world_demonstrations}
    \small
    \setlength{\tabcolsep}{0pt}
    \begin{tabular}{@{}>{\raggedright\arraybackslash}p{0.44\columnwidth}>{\centering\arraybackslash}p{0.28\columnwidth}>{\centering\arraybackslash}p{0.28\columnwidth}@{}}
        \toprule
        Task & Demonstrations & Frames \\
        \midrule
        Find Block & 50 & 33,660 \\
        Sequential Grasp & 50 & 40,004 \\
        Find Heaviest & 50 & 57,073 \\
        Press Button & 50 & 13,285 \\
        \bottomrule
    \end{tabular}
    \begin{tablenotes}[flushleft]
        \footnotesize
        \item[] \textit{Note.} All datasets are sampled at 15\,Hz.
    \end{tablenotes}
    \end{threeparttable}
\end{table}

\paragraph{\texttt{Find Block}.}
A block is hidden beneath one of three inverted cups.
The robot must lift and inspect each cup from left to right, replacing it before moving to the next.
After inspecting all three cups, it must move the cup covering the block forward, then grasp and lift the block.
Replacing the cups conceals their contents again, so the final retrieval requires remembering which cup covered the block.

\paragraph{\texttt{Sequential Grasp}.}
The robot must place a green cuboid, a red cylinder, and an orange cube on a designated tabletop sticker in that order.
Each object must be returned to its original position before the next is moved; after all three placement-and-return cycles, the robot presses the completion button.
Because each cycle restores the object layout, the policy must track completed cycles to select the next object and recognize when to finish.

\paragraph{\texttt{Find Heaviest}.}
Three identical-looking blocks must be weighed from left to right, with each returned to its original position after its weight is read.
The robot must then place the heaviest block on the scale.
This requires associating earlier weight readings with block locations and comparing the measurements while tracking which blocks have been weighed.

\paragraph{\texttt{Press Button}.}
A number displayed on paper specifies a target count $N\in\{1,\ldots,5\}$.
The robot must press the left button exactly $N$ times, then press the right button once to finish.
The policy must track cumulative left-button activations across repeated presses and releases and use the target count to decide when to issue the final right-button press.

Evaluation settings and results are provided in Appendix~\ref{sec:real_world_evaluation_details}.
Recorded task sequences are shown in Appendix~\ref{sec:real_world_rollouts}.

\subsection{Additional Experimental Details}
\label{sec:additional_experimental_details}

\subsubsection{RMBench Evaluation Details}
\label{sec:rmbench_additional_details}

Table~\ref{tab:rmbench_full} reports nine-task results, with baselines from EventVLA~\cite{yang2026eventvla}.
Both KeyMemVLA policies are jointly trained on all nine tasks and retain SPH.
Each policy is evaluated over 100 trials per task.
The full policy uses KPH and keyframe memory entries; w/o KPH removes keyframe inputs, prediction, supervision, and writes.
Both use the initial observation and offsets $-30$, $-15$, and $0$: historical observations retain the main view, while current observations retain all three views.
EventVLA retains all three views at every anchor.

\begin{table*}[!tbp]
    \centering
    \begin{threeparttable}
    \caption{\textbf{Per-task success rates on RMBench.}}
    \label{tab:rmbench_full}
    \footnotesize
    \setlength{\tabcolsep}{2.5pt}
    \renewcommand{\arraystretch}{0.94}
    \begin{tabular*}{\textwidth}{@{\extracolsep{\fill}}lcccccccccc@{}}
        \toprule
        & \multicolumn{5}{c}{$M(1)$} & \multicolumn{4}{c}{$M(n)$} & \\
        \cmidrule(lr){2-6}\cmidrule(lr){7-10}
        Method & \shortstack{Observe and\\Pickup} & \shortstack{Rearrange\\Blocks}
        & \shortstack{Put Back\\Block} & \shortstack{Swap\\Blocks} & Swap T
        & \shortstack{Battery\\Try} & \shortstack{Blocks\\Ranking Try}
        & \shortstack{Cover\\Blocks} & \shortstack{Press\\Button} & Average \\
        \midrule
        \rowcolor{kmGroupShade}
        \multicolumn{11}{l}{$\blacktriangledown$ \textit{Non-Memory-Based Policies}} \\
        ACT~\cite{zhao2023act} & 1\% & 29\% & 0\% & 2\% & 2\% & 19\% & 0\% & 0\% & 0\% & 5.90\% \\
        DP~\cite{chi2023diffusion} & 1\% & 0\% & 0\% & 11\% & 20\% & 10\% & 10\% & 0\% & 0\% & 5.80\% \\
        $\pi_{0.5}$~\cite{intelligence2025pi05} & 9\% & 13\% & 11\% & 24\% & 15\% & 16\% & 6\% & 0\% & 0\% & 10.40\% \\
        Qwen-OFT~\cite{kim2025oft} & 0\% & 0\% & 0\% & 0\% & 0\% & 14\% & 37\% & 0\% & 0\% & 5.60\% \\
        X-VLA~\cite{zheng2026xvla} & 9\% & 13\% & 18\% & 16\% & 3\% & 26\% & 1\% & 2\% & 0\% & 9.80\% \\
        \rowcolor{kmGroupShade}
        \multicolumn{11}{l}{$\blacktriangledown$ \textit{Dual-System Memory-Based Vision-Language-Action Models}} \\
        MemER~\cite{sridhar2025memer} & 7\% & 17\% & 0\% & 14\% & 7\% & 27\% & 0\% & 6\% & 0\% & 8.70\% \\
        Mem-0~\cite{chen2026rmbench} & 4\% & \kmSecond{89\%} & 90\% & 67\% & 14\% & 28\% & 18\% & 68\% & 0\% & 42.00\% \\
        \rowcolor{kmGroupShade}
        \multicolumn{11}{l}{$\blacktriangledown$ \textit{End-to-End Memory-Based Vision-Language-Action Models}} \\
        MemoryVLA (OpenVLA)~\cite{shi2026memoryvla} & 0\% & 22\% & 50\% & 17\% & 9\% & 25\% & 12\% & 40\% & 0\% & 19.40\% \\
        MemoryVLA (Qwen-OFT)~\cite{shi2026memoryvla} & 2\% & 53\% & 81\% & 76\% & 9\% & 33\% & 53\% & 69\% & 0\% & 41.70\% \\
        EventVLA (VA only)~\cite{yang2026eventvla} & 21\% & \kmBest{96\%} & \kmSecond{95\%} & \kmBest{96\%} & \kmSecond{87\%} & 35\% & \kmSecond{81\%} & \kmSecond{97\%} & 3\% & \kmSecond{67.80\%} \\
        KeyMemVLA w/o KPH & \kmSecond{22\%} & \kmSecond{89\%} & 76\% & \kmSecond{91\%} & 86\% & \kmBest{42\%} & \kmBest{100\%} & \kmBest{100\%} & \kmSecond{8\%} & \kmBest{68.22\%} \\
        KeyMemVLA & \kmBest{28\%} & 84\% & \kmBest{100\%} & 80\% & \kmBest{91\%} & \kmSecond{36\%} & 66\% & 70\% & \kmBest{26\%} & 64.56\% \\
        \bottomrule
    \end{tabular*}
    \begin{tablenotes}[flushleft]
        \footnotesize
        \item[] \textit{Note.} Baseline task rates and aggregate averages follow EventVLA~\cite{yang2026eventvla}; its RMBench row uses VA only.
        Both KeyMemVLA policies jointly train on all nine tasks and retain SPH; w/o KPH removes keyframe prediction and memory.
        Shaded bold and underlined entries denote the best and second-best distinct results per column, including ties.
    \end{tablenotes}
    \end{threeparttable}
\end{table*}

\paragraph{Task-level analysis.}
Under joint nine-task training, adding KPH and keyframe memory improves some RMBench tasks but lowers overall success.
Initial and recent observations may already provide sufficient cues for many tasks, so additional keyframes may introduce redundancy and make relevant evidence harder to use.
Without KPH, KeyMemVLA performs comparably to EventVLA (VA only) overall, with higher success on all four $M(n)$ tasks but a lower average success rate on $M(1)$ tasks.
Explicit current-subtask prediction may help track repeated stages, such as successive presses in \texttt{Press Button}, whereas EventVLA's richer historical multi-view context may better support spatial decisions in $M(1)$ tasks.
These observations motivate KMBench's focus on retaining evidence acquired during intermediate interactions for later decisions across subtasks.

\subsubsection{Real-World Evaluation Details}
\label{sec:real_world_evaluation_details}

\paragraph{Platform and Control.}
A Flexiv Rizon4s with a Flexiv-GN01 gripper uses left-main, right-auxiliary, and wrist RGB views.
Cameras capture $640\times480$ images at 30\,Hz, resized to $256\times256$.
Each action comprises a 3D translation increment, a 6D rotation-increment representation, and an absolute gripper-width target.
Translation increments use the robot base frame; outputs are unnormalized using the checkpoint's training statistics and converted into robot targets.
The nominal action rate is 15\,Hz, with local interpolation at 100\,Hz.
Inference uses one NVIDIA A100 GPU and the procedure in Appendix~\ref{sec:real_world_deployment}.

\paragraph{Evaluation Protocol.}
KeyMemVLA, $\pi_{0.5}$, and MemER are each jointly trained on all four real-world tasks.
Each method is evaluated over 10 trials per task.
For \texttt{Press Button}, each target count has two test trials, ordered from $N=1$ to $N=5$.

\paragraph{Task and Subtask Success Rates.}
Let $c_i$ denote the number of successfully completed subtasks in trial $i$, and $K_i$ the number required.
For $R=10$ trials per task and method, task success rate (TSR) and subtask success rate (SSR) are
\begin{equation}
    \mathrm{TSR}=\frac{1}{R}\sum_{i=1}^{R}\mathbf{1}[c_i=K_i],
    \qquad
    \mathrm{SSR}=\frac{\sum_{i=1}^{R}c_i}{\sum_{i=1}^{R}K_i}.
    \label{eq:real_world_success}
\end{equation}
\texttt{Find Block} has five subtasks: three cup inspections, moving the target cup, and lifting the block.
\texttt{Sequential Grasp} has seven: three placement/return pairs and the completion-button activation.
\texttt{Find Heaviest} has seven: three weighing/return pairs and the final placement of the heaviest block.
For \texttt{Press Button}, the $N_i$ left-button activations and final right-button activation give $K_i=N_i+1$.
The resulting required subtask counts $K_i$ across the 10 trials are $(2,2,3,3,4,4,5,5,6,6)$.
The SSR denominators are consequently 50, 70, 70, and 40 for the four tasks, respectively.
The overall TSR is the unweighted average of the four task rates.

Table~\ref{tab:real_world_subtask_success} reports task and subtask success rates.

\begin{table*}[!tbp]
    \centering
    \begin{threeparttable}
    \caption{\textbf{Real-world task success rate (TSR) and subtask success rate (SSR).}}
    \label{tab:real_world_subtask_success}
    \small
    \setlength{\tabcolsep}{3pt}
    \begin{tabular*}{\textwidth}{@{\extracolsep{\fill}}l*{8}{c}@{}}
        \toprule
        Method & \multicolumn{2}{c}{Find Block} & \multicolumn{2}{c}{Sequential Grasp}
        & \multicolumn{2}{c}{Find Heaviest} & \multicolumn{2}{c}{Press Button} \\
        \cmidrule(lr){2-3}\cmidrule(lr){4-5}\cmidrule(lr){6-7}\cmidrule(lr){8-9}
        & TSR & SSR & TSR & SSR & TSR & SSR & TSR & SSR \\
        \midrule
        $\pi_{0.5}$ & 0.0\% & 14.0\% & 0.0\% & 4.3\% & 0.0\% & 8.6\% & 10.0\% & \kmSecond{62.5\%} \\
        MemER & \kmSecond{20.0\%} & \kmSecond{40.0\%} & \kmSecond{40.0\%} & \kmBest{88.6\%} & \kmSecond{20.0\%} & \kmSecond{60.0\%} & \kmSecond{20.0\%} & 45.0\% \\
        \textbf{KeyMemVLA} & \kmBest{50.0\%} & \kmBest{68.0\%} & \kmBest{50.0\%} & \kmSecond{70.0\%} & \kmBest{30.0\%} & \kmBest{61.4\%} & \kmBest{60.0\%} & \kmBest{90.0\%} \\
        \bottomrule
    \end{tabular*}
    \begin{tablenotes}[flushleft]
        \footnotesize
        \item[] \textit{Note.} Each method uses 10 trials per task. SSR pools completed subtasks over all required subtasks.
        Shaded bold and underlined values mark the best and second-best results in each column.
    \end{tablenotes}
    \end{threeparttable}
\end{table*}

\paragraph{Inference Efficiency.}
To complement the task-success evaluation in Sec.~\ref{sec:real_world}, we measure inference efficiency on \texttt{Sequential Grasp} using a single NVIDIA A100 GPU.
For each method ($\pi_{0.5}$, MemER, and KeyMemVLA), we perform three complete inference runs, each comprising multiple action-chunk inference calls.

Let $n_r$ denote the number of inference calls in run $r$, and let $t_{r,i}$ and $m_{r,i}$ denote the latency and GPU memory usage recorded for its $i$-th call.
We first compute the arithmetic averages within each run:
\begin{equation}
    \bar t_r=\frac{1}{n_r}\sum_{i=1}^{n_r}t_{r,i},
    \qquad
    \bar m_r=\frac{1}{n_r}\sum_{i=1}^{n_r}m_{r,i}.
    \label{eq:efficiency_run_means}
\end{equation}
The reported mean inference latency and mean GPU memory usage are then obtained by averaging the three run-level results:
\begin{equation}
    \bar t=\frac{1}{3}\sum_{r=1}^{3}\bar t_r,
    \qquad
    \bar m=\frac{1}{3}\sum_{r=1}^{3}\bar m_r.
    \label{eq:efficiency_overall_means}
\end{equation}
Each complete run therefore contributes equally to the reported means.
Table~\ref{tab:real_world_runtime} reports these averages together with device-wide GPU utilization.

\begin{table}[!htbp]
    \centering
    \begin{threeparttable}
    \caption{\textbf{Real-world inference efficiency on a single NVIDIA A100 GPU.}}
    \label{tab:real_world_runtime}
    \small
    \setlength{\tabcolsep}{2pt}
    \begin{tabular*}{\columnwidth}{@{\extracolsep{\fill}}lccc@{}}
        \toprule
        Method & \shortstack{Latency\\(ms)} & \shortstack{GPU memory\\(GiB)} & GPU util. \\
        \midrule
        $\pi_{0.5}$ & 276 & 7.73 & 28.5\% \\
        MemER & 1432 & 17.05 & 32.4\% \\
        KeyMemVLA & 640 & 10.20 & 61.8\% \\
        \bottomrule
    \end{tabular*}
    \begin{tablenotes}[flushleft]
        \footnotesize
        \item[] \textit{Note.} All values are means; utilization is measured for the whole GPU.
    \end{tablenotes}
    \end{threeparttable}
\end{table}

\subsection{Ablation and Linear-Probe Details}
\label{sec:ablation_probe_details}

\paragraph{Policy ablations.}
All variants share data, initialization, training budget, visual anchors, and task instructions.
Removing KPH removes keyframe supervision, prediction, writes, and inputs; removing SPH removes subtask supervision, prediction, and text conditioning.
Table~\ref{tab:component_ablations} summarizes average success in the main paper; Tables~\ref{tab:ablation_per_task} and~\ref{tab:kmbench_ablation_per_task} provide the corresponding per-task results on RoboTwin-MeM and KMBench.

\begin{table}[!htbp]
    \centering
    \begin{threeparttable}
    \caption{\textbf{Per-task ablation success rates on RoboTwin-MeM.}}
    \label{tab:ablation_per_task}
    \small
    \setlength{\tabcolsep}{2pt}
    \renewcommand{\arraystretch}{0.96}
    \begin{tabular*}{\columnwidth}{@{\extracolsep{\fill}}lccc@{}}
        \toprule
        Task & Full & w/o KPH & w/o SPH \\
        \midrule
        Cover Blocks Hard & \kmSecond{91\%} & 6\% & \kmBest{94\%} \\
        Find Seal and Seal Stamp & \kmBest{80\%} & \kmSecond{69\%} & 48\% \\
        Pick Objects in Order & \kmBest{100\%} & 11\% & \kmSecond{72\%} \\
        Pick the Unhidden Block & \kmBest{100\%} & 27\% & \kmSecond{43\%} \\
        Press Button Keyframe & \kmBest{74\%} & 30\% & \kmSecond{61\%} \\
        Put Back Block Hard & \kmBest{100\%} & \kmSecond{38\%} & \kmBest{100\%} \\
        Rearrange Blocks Hard & \kmBest{96\%} & \kmSecond{72\%} & 52\% \\
        Reproduce Route & \kmBest{100\%} & 0\% & \kmSecond{27\%} \\
        \midrule
        Average & \kmBest{92.63\%} & 31.63\% & \kmSecond{62.13\%} \\
        \bottomrule
    \end{tabular*}
    \begin{tablenotes}[flushleft]
        \footnotesize
        \item[] \textit{Note.} Shaded bold and underlined values mark the best and second-best distinct results within each row, including ties.
    \end{tablenotes}
    \end{threeparttable}
\end{table}

\begin{table}[!htbp]
    \centering
    \begin{threeparttable}
    \caption{\textbf{Per-task ablation success rates on KMBench.}}
    \label{tab:kmbench_ablation_per_task}
    \small
    \setlength{\tabcolsep}{2pt}
    \renewcommand{\arraystretch}{0.96}
    \begin{tabular*}{\columnwidth}{@{\extracolsep{\fill}}lccc@{}}
        \toprule
        Task & Full & w/o KPH & w/o SPH \\
        \midrule
        Weight Rank & \kmSecond{26\%} & \kmBest{35\%} & 19\% \\
        Bin Color Query & \kmSecond{90\%} & 0\% & \kmBest{100\%} \\
        Scan and Route & \kmBest{48\%} & 17\% & \kmSecond{33\%} \\
        Button Pair & \kmBest{97\%} & 17\% & \kmSecond{83\%} \\
        Two-Bin Count & \kmBest{97\%} & 38\% & \kmSecond{76\%} \\
        Conveyor Color Count & \kmSecond{57\%} & 11\% & \kmBest{60\%} \\
        Calibrated Sort & \kmBest{45\%} & 36\% & \kmSecond{44\%} \\
        Battery Charge & \kmSecond{91\%} & 20\% & \kmBest{95\%} \\
        \midrule
        Average & \kmBest{68.88\%} & 21.75\% & \kmSecond{63.75\%} \\
        \bottomrule
    \end{tabular*}
    \begin{tablenotes}[flushleft]
        \footnotesize
        \item[] \textit{Note.} Shaded bold and underlined values mark the best and second-best distinct results within each row, including ties.
    \end{tablenotes}
    \end{threeparttable}
\end{table}

\paragraph{Frozen representations.}
The probes assess the linear decodability of temporal progress from action-query representations on RoboTwin-MeM.
We freeze the final full-policy and \emph{w/o SPH} checkpoints in evaluation mode, using paired instructions, current and initial observations, recent history, and fixed visible keyframes.
Online memory writes are disabled.
The full policy first predicts the current subtask and conditions its action-prediction pass on the predicted text; \emph{w/o SPH} uses its native prompt without subtask text.
We mean-pool the 50 action-query representations from each model's action-prediction pass into $h$, taking features from the final decoder block before the final RMSNorm.

\paragraph{Probe training.}
Each task contributes 40 probe-training and 10 probe-test episodes, giving 320/80 episodes across eight tasks.
Training uses 64 preselected timesteps per trajectory, one randomly drawn per temporal interval; testing uses 128 uniformly spaced timesteps including both endpoints.
This yields 20,480 training and 10,240 test samples per checkpoint.
Episodes and sample times are paired across models.
Targets are normalized frame progress, $p=t/(T-1)$.
Each checkpoint has one linear probe shared across tasks, with features standardized using training statistics.
For standardized features $\widetilde h$, the probe and objective are
\begin{equation}
    \begin{aligned}
        \hat p_i&=w^\top\widetilde h_i+b,\\
        \mathcal L_{\mathrm{probe}}
        &=\frac{1}{N_{\mathrm{train}}}\sum_i|\hat p_i-p_i|.
    \end{aligned}
    \label{eq:progress_probe_training}
\end{equation}
Both probes use Adam, learning rate $10^{-3}$, batch size 64, 300 epochs, and seed 42.
All 400 episodes were included in training both policies; the 320/80 split applies only to probe fitting and evaluation.

\paragraph{Evaluation.}
Test MAE averages absolute errors equally over sampled times, episodes within each task, and finally tasks, using raw predictions without clipping, smoothing, or monotonic constraints.
Table~\ref{tab:sph_probe_per_task} reports per-task MAEs and their equal-weight averages across the eight tasks.
Relative reduction is computed from the unrounded averages of the eight task MAEs as
$r=(\mathrm{MAE}_{\mathrm{w/o\ SPH}}-\mathrm{MAE}_{\mathrm{w/\ SPH}})/\mathrm{MAE}_{\mathrm{w/o\ SPH}}\times100\%$.
The relative reduction is 23.12\%.

\begin{table}[!htbp]
    \centering
    \begin{threeparttable}
    \caption{\textbf{Linear-probe test MAE by task on RoboTwin-MeM.}}
    \label{tab:sph_probe_per_task}
    \small
    \setlength{\tabcolsep}{1.5pt}
    \renewcommand{\arraystretch}{0.96}
    \begin{tabular*}{\columnwidth}{@{\extracolsep{\fill}}lccc@{}}
        \toprule
        Task & \shortstack{w/o SPH\\MAE} & \shortstack{w/ SPH\\MAE} & Reduction \\
        \midrule
        Cover Blocks Hard & 0.09496 & \textbf{0.08621} & 9.21\% \\
        Find Seal and Seal Stamp & 0.13256 & \textbf{0.09550} & 27.96\% \\
        Pick Objects in Order & 0.10476 & \textbf{0.09554} & 8.80\% \\
        Pick the Unhidden Block & 0.08758 & \textbf{0.07261} & 17.10\% \\
        Press Button Keyframe & 0.20162 & \textbf{0.06063} & 69.93\% \\
        Put Back Block Hard & \textbf{0.08616} & 0.10324 & $-19.82\%$ \\
        Rearrange Blocks Hard & 0.09451 & \textbf{0.08054} & 14.78\% \\
        Reproduce Route & 0.11649 & \textbf{0.11199} & 3.86\% \\
        \midrule
        Average & 0.11483 & \textbf{0.08828} & 23.12\% \\
        \bottomrule
    \end{tabular*}
    \begin{tablenotes}[flushleft]
        \footnotesize
        \item[] \textit{Note.} Lower MAE is bold; reduction is relative to w/o SPH, with negative values indicating increased error.
        MAEs and reductions are rounded independently from unrounded predictions.
    \end{tablenotes}
    \end{threeparttable}
\end{table}

\subsection{Key Memory Mode Details}
\label{sec:memory_mode_details}

The three modes use the same RoboTwin-MeM demonstrations, backbone, objectives, visual anchors, SPH, and KPH.
Each is jointly trained on all eight tasks.
\emph{Image} stores keyframe images, \emph{Text} stores key-subtask descriptions, and \emph{ImageAndText} stores both (Appendix~\ref{sec:vlm_prompt_design}).

Training uses annotated key-subtask endpoints and their descriptions.
At inference, the predicted current subtask is bound to a proposed write time.
An accepted event commits the acquired image, the bound description, or both when execution reaches that time.
Image--text pairs share a timestamp and remain paired through eviction.
Descriptions at distinct events are retained even when identical.

Memory starts empty and retains at most five events in acquisition order, evicting the oldest when full. All modes use only events with $t'<t$.
Table~\ref{tab:memory_modality_per_task} gives the per-task results and equal-weight averages.

\begin{table}[!htbp]
    \centering
    \begin{threeparttable}
    \caption{\textbf{Per-task success rates by key memory mode on RoboTwin-MeM.}}
    \label{tab:memory_modality_per_task}
    \small
    \setlength{\tabcolsep}{2pt}
    \renewcommand{\arraystretch}{0.96}
    \begin{tabular*}{\columnwidth}{@{\extracolsep{\fill}}lccc@{}}
        \toprule
        Task & \emph{Image} & \emph{ImageAndText} & \emph{Text} \\
        \midrule
        Cover Blocks Hard & \kmBest{91\%} & \kmSecond{79\%} & 5\% \\
        Find Seal and Seal Stamp & \kmBest{80\%} & \kmSecond{60\%} & 35\% \\
        Pick Objects in Order & \kmBest{100\%} & 33\% & \kmSecond{36\%} \\
        Pick the Unhidden Block & \kmBest{100\%} & \kmBest{100\%} & \kmSecond{26\%} \\
        Press Button Keyframe & \kmBest{74\%} & 28\% & \kmSecond{39\%} \\
        Put Back Block Hard & \kmBest{100\%} & \kmBest{100\%} & \kmBest{100\%} \\
        Rearrange Blocks Hard & \kmBest{96\%} & \kmSecond{95\%} & 90\% \\
        Reproduce Route & \kmBest{100\%} & \kmSecond{83\%} & \kmBest{100\%} \\
        \midrule
        Average & \kmBest{92.63\%} & \kmSecond{72.25\%} & 53.88\% \\
        \bottomrule
    \end{tabular*}
    \begin{tablenotes}[flushleft]
        \footnotesize
        \item[] \textit{Note.} Shaded bold and underlined values mark the best and second-best distinct results within each row, including ties.
    \end{tablenotes}
    \end{threeparttable}
\end{table}

\newpage
\kmrolloutfloatsafterheading
\balance
\section{Task Rollouts}
\label{sec:task_rollouts}

We present recorded policy rollouts on KMBench and the real robot.
Each panel contains 12 observations in chronological order, read left to right across the top row and then the bottom row.

\subsection{KMBench Rollouts}
\label{sec:kmbench_rollouts}

\begin{figure*}
    \centering
    \includegraphics[width=\textwidth]{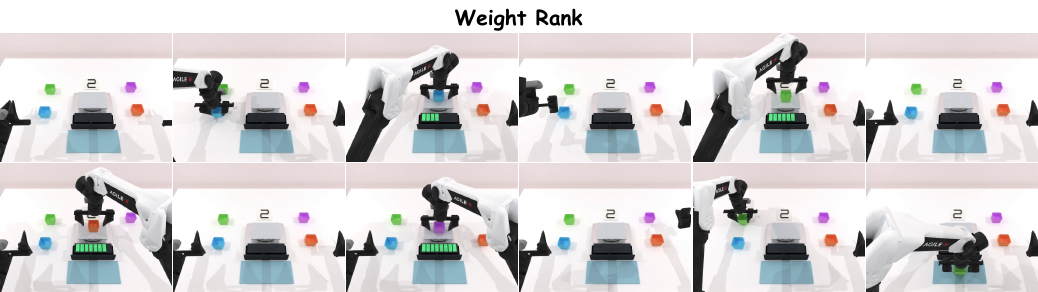}
    \caption{\textbf{Weight Rank rollout.} The displayed target rank is 2. The robot weighs the four boxes, returns them to their original locations, and places the second-lightest box, shown in green, on the central pad.}
    \label{fig:rollout_weight_rank}
\end{figure*}

\begin{figure*}
    \centering
    \includegraphics[width=\textwidth]{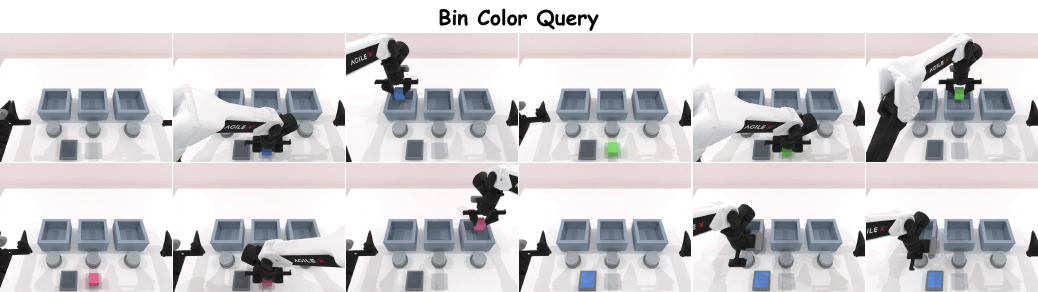}
    \caption{\textbf{Bin Color Query rollout.} The robot deposits the blue, green, and pink blocks into the left, middle, and right bins, respectively. After the blocks disappear into the bins, it answers the blue query by pressing the left button.}
    \label{fig:rollout_bin_color_query}
\end{figure*}

\begin{figure*}
    \centering
    \includegraphics[width=\textwidth]{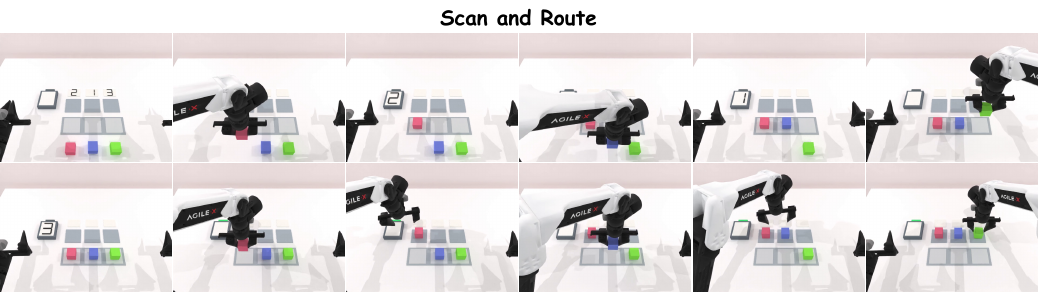}
    \caption{\textbf{Scan and Route rollout.} The initial station identifiers are 2, 1, and 3 from left to right. Registration assigns the pink, blue, and green items to stations 2, 1, and 3, respectively. After the readouts clear, the robot delivers the items to the corresponding stations.}
    \label{fig:rollout_scan_and_route}
\end{figure*}

\begin{figure*}
    \centering
    \includegraphics[width=\textwidth]{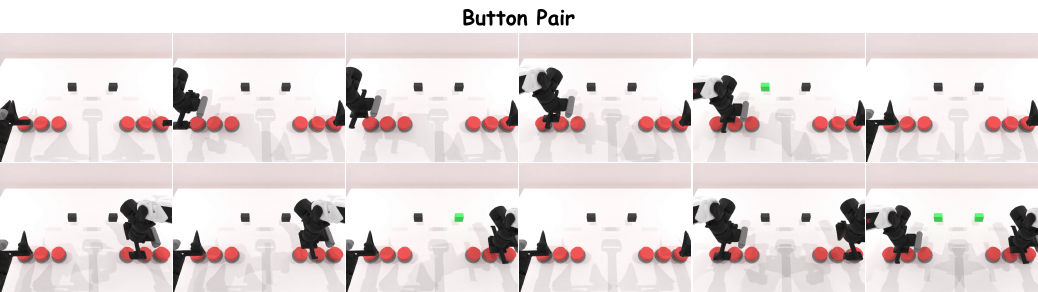}
    \caption{\textbf{Button Pair rollout.} The robot tests candidates on each side and observes the light feedback. It releases the buttons and resets the arms after each side's exploration, then simultaneously presses the middle button on each side. Both indicators light up in the last frame.}
    \label{fig:rollout_button_pair}
\end{figure*}

\begin{figure*}
    \centering
    \includegraphics[width=\textwidth]{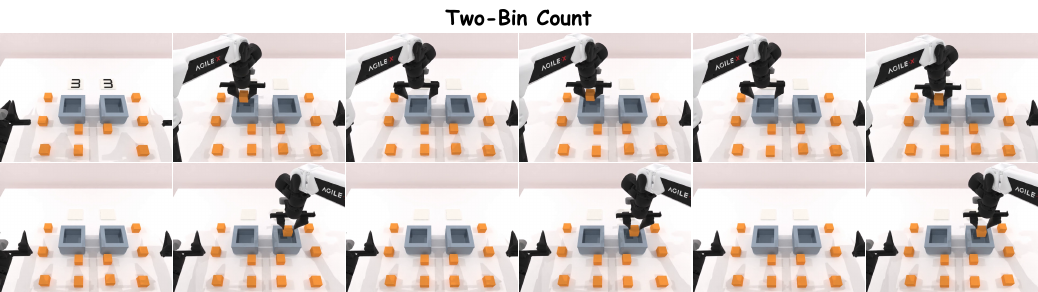}
    \caption{\textbf{Two-Bin Count rollout.} The initial targets are three deposits into each bin. The robot fills the left bin first, then the right. Deposited blocks disappear and the stock is replenished, so the robot must track completed transfers. The last frame shows the final deposit.}
    \label{fig:rollout_two_bin_count}
\end{figure*}

\begin{figure*}
    \centering
    \includegraphics[width=\textwidth]{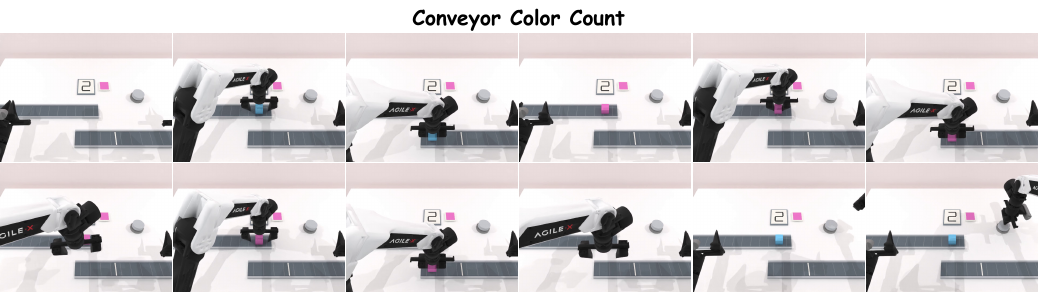}
    \caption{\textbf{Conveyor Color Count rollout.} The target is two pink-block exits. The robot transfers a blue block followed by two pink blocks to the output conveyor. After the second pink block exits, it presses the completion button, leaving the next blue block on the input conveyor.}
    \label{fig:rollout_conveyor_color_count}
\end{figure*}

\begin{figure*}
    \centering
    \includegraphics[width=\textwidth]{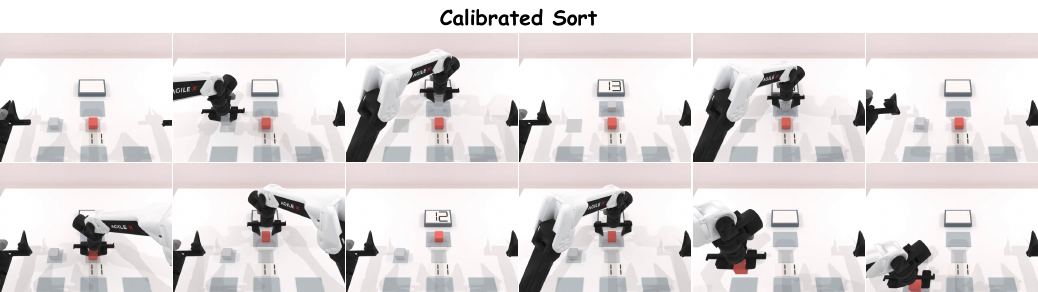}
    \caption{\textbf{Calibrated Sort rollout.} Reference and test readouts are 13 and 12, with a threshold of 11. Calibration gives $v=12-13+10=9<11$, so the test block is placed on the left pad. The readouts clear after object removal.}
    \label{fig:rollout_calibrated_sort}
\end{figure*}

\begin{figure*}
    \centering
    \includegraphics[width=\textwidth]{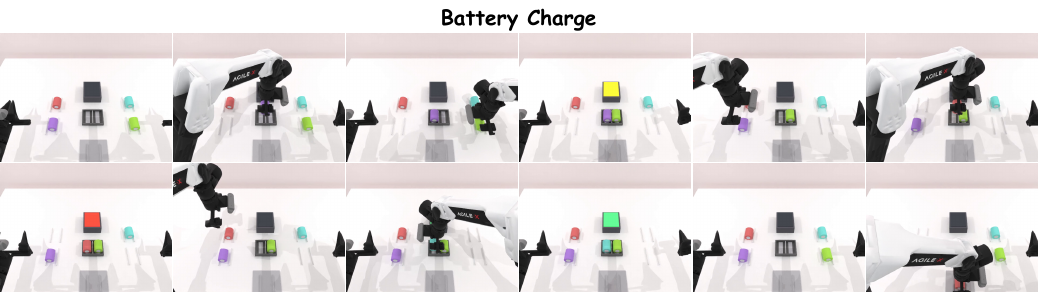}
    \caption{\textbf{Battery Charge rollout.} The green battery remains in the right slot while the purple, red, and cyan batteries are tested in the left slot, producing yellow, red, and green feedback, respectively. The robot returns the tested batteries and places the lowest-charge battery, shown in red, on the central pad.}
    \label{fig:rollout_battery_charge}
\end{figure*}

Figures~\ref{fig:rollout_weight_rank}--\ref{fig:rollout_battery_charge} show main-view observations from one successful KeyMemVLA policy rollout per KMBench task.
Task definitions are given in Appendix~\ref{sec:kmbench_additional_details}.
The selected observations show object grasps, transfers, and placements, together with initial task cues, key feedback, and final decisions.

\subsection{Real-World Rollouts}
\label{sec:real_world_rollouts}

\begin{figure*}
    \centering
    \includegraphics[width=\textwidth]{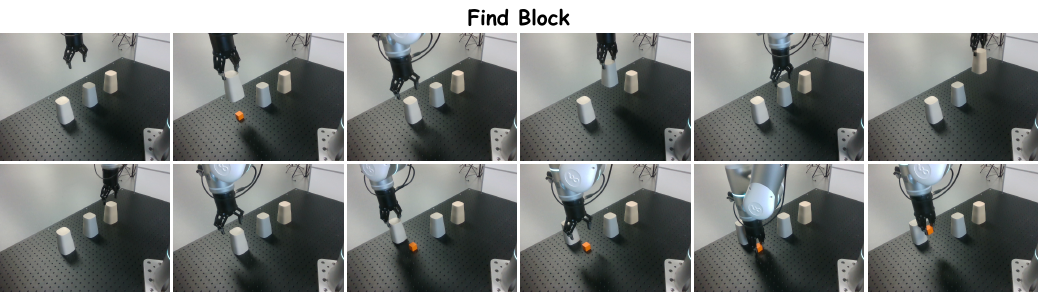}
    \caption{\textbf{Find Block rollout.} Sequential cup inspection is followed by revisiting the cup covering the orange block, moving that cup, and grasping and lifting the block.}
    \label{fig:rollout_find_block}
\end{figure*}

Figures~\ref{fig:rollout_find_block}--\ref{fig:rollout_press_button} show recorded executions of the four tasks defined in Appendix~\ref{sec:real_world_details}.
Aggregate success rates are reported in Table~\ref{tab:real_world_subtask_success}.

\begin{figure*}
    \centering
    \includegraphics[width=\textwidth]{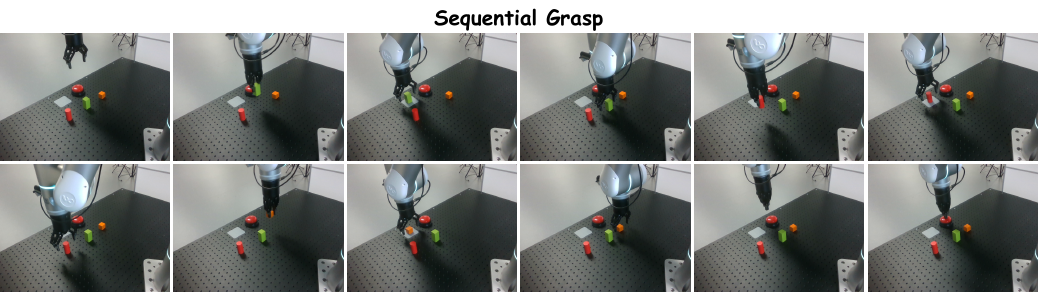}
    \caption{\textbf{Sequential Grasp rollout.} The sequence shows placement and return of the green, red, and orange objects in order, followed by the completion-button interaction.}
    \label{fig:rollout_sequential_grasp}
\end{figure*}

\begin{figure*}
    \centering
    \includegraphics[width=\textwidth]{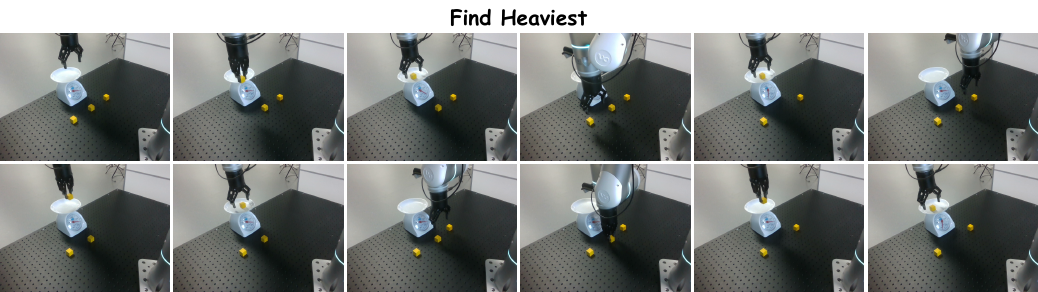}
    \caption{\textbf{Find Heaviest rollout.} Successive weighing and return operations are followed by picking a previously weighed block and placing it on the scale.}
    \label{fig:rollout_find_heaviest}
\end{figure*}

\begin{figure*}
    \centering
    \includegraphics[width=\textwidth]{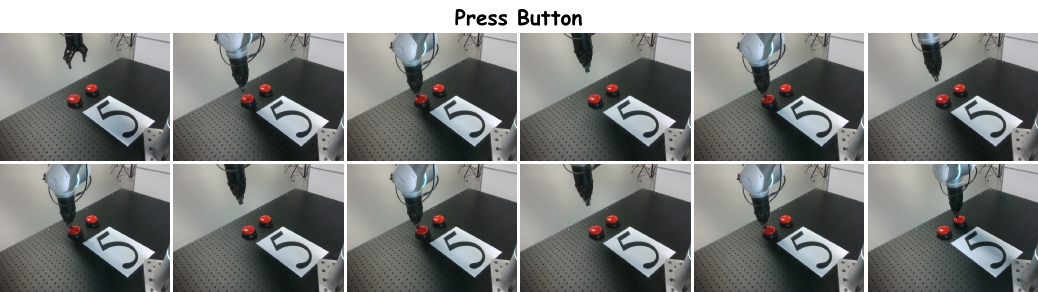}
    \caption{\textbf{Press Button rollout.} The visible target is $N=5$. Repeated left-button interactions are interleaved with release motions, followed by the right-button interaction in the last frame.}
    \label{fig:rollout_press_button}
\end{figure*}

\end{document}